\documentclass[letterpaper, 10 pt, conference]{ieeeconf}  % Comment this line out if you need a4paper

\IEEEoverridecommandlockouts                              % This command is only needed if 
\usepackage{amsmath} % assumes amsmath package installed
\usepackage{gensymb}
\usepackage{textcomp}
\usepackage{mathtools}
\usepackage{xargs}  
\usepackage{units}
\usepackage{pgfplots}
\pgfplotsset{compat=1.6}
\usepackage{caption}
\usepackage{subcaption}
\usepackage{graphicx,xcolor} 
\usepackage{colortbl}
\usepackage{color}
\usepackage{multicol} 
\usepackage{cprotect}
\usepackage{placeins}
\usepackage{balance}
\usepackage{siunitx}
\usepackage{verbatim}
\usepackage{graphicx,subcaption}
\usepackage{float}
\usepackage[font=small,labelfont=bf,tableposition=top]{caption}

\usepackage[bordercolor=white,backgroundcolor=gray!30,linecolor=black,colorinlistoftodos]{todonotes}
\newcommandx{\timo}[1]{\todo[color=yellow, inline]{timo: {#1}}}
\newcommandx{\igor}[1]{\todo[color=green, inline]{igor: {#1}}}
\newcommandx{\cesar}[1]{\todo[color=lightgray, inline]{cesar: {#1}}}
\newcommandx{\tj}[1]{\todo[color=red, inline]{tj: {#1}}}

\makeatletter
\let\NAT@parse\undefined
\makeatother

\usepackage{nameref,zref-user, zref-xr}
\usepackage{hyperref}
\zexternaldocument{letter}

\title{\LARGE \bf
Free LSD: Prior-Free Visual Landing Site Detection\\for Autonomous Planes
}

\author{Timo Hinzmann$^{1}$, Thomas Stastny$^{1}$, Cesar Cadena$^{1}$, Roland Siegwart$^{1}$, and Igor Gilitschenski$^{1,2}$ %\\ \\% <-this % stops a space
\thanks{$^{1}$Autonomous Systems Lab, ETH Zurich, Switzerland.}%
\thanks{$^{2}$CSAIL, MIT, Cambridge, MA, USA.}
}
\usepackage{xcolor}

\DeclareCaptionLabelFormat{andtable}{#1~#2  \&  \tablename~\thetable}

\begin{document}

\maketitle
\thispagestyle{empty}
\pagestyle{empty}

\newcommand{\IntroFirstWord}{Small-scale }

%%%%%%%%%%%%%%%%%%%%%%%%%%%%%%%%%%%%%%%
%% SWITCH: REVIEW TAGS (ON/OFF)
%%%%%%%%%%%%%%%%%%%%%%%%%%%%%%%%%%%%%%%
\newif\ifReviewTags
%\ReviewTagstrue % Defined? --> review tags
\ifReviewTags
\newcommand{\review}[2]{{\color{blue} #1 ({\zref{#2}})}}
\newcommand{\reviews}[3]{{\color{blue} #1 ({\zref{#2}}, \zref{#3})}}
\else
\newcommand{\review}[2]{#1}
\newcommand{\reviews}[3]{#1}
\fi
%
%%%%%%%%%%%%%%%%%%%%%%%%%%%%%%%%%%%%%%%
%% SWITCH: IMAGE RESOLUTION (High/Low)
%%%%%%%%%%%%%%%%%%%%%%%%%%%%%%%%%%%%%%%
\newif\ifMaxQuality
%\MaxQualitytrue % Defined? --> max. quality
\newcommand{\includegraphicsHighLowRes}[4]{
\ifMaxQuality \includegraphics[#1]{#2}\caption[]{#4} \else \includegraphics[#1]{#3}\caption[]{#4} \fi}

%%%%%%%%%%%%%%%%%%%%%%%%%%%%%%%%%%%%%%%%%%%%%%%%%%%%%%%%%%%%%%%%%%%%%%%%%%%%%%%%
\begin{abstract}
Full autonomy for fixed-wing unmanned aerial vehicles (UAVs) requires the capability to autonomously detect potential landing sites in unknown and unstructured terrain, allowing for self-governed mission completion or handling of emergency situations.
In this work, we propose a \review{perception system}{review1:conclusion} addressing this challenge by detecting landing sites based on their texture and geometric shape without using any prior knowledge about the environment.
The proposed method considers hazards within the landing region such as terrain roughness and slope, surrounding obstacles that obscure the landing approach path, and the local wind field that is estimated by the on-board EKF.
The latter enables applicability of the proposed method on small-scale autonomous planes without landing gear.
%
% REMARK(ig): Removed the word "planned" here if TJ wants to go for a follow-up paper.
A safe approach path is computed based on the UAV dynamics, expected state estimation and actuator uncertainty, and the on-board computed elevation map.
The proposed framework has been successfully tested on photo-realistic synthetic datasets and in challenging real-world environments.
%.
\end{abstract}

%%%%%%%%%%%%%%%%%%%%%%%%%%%%%%%%%%%%%%%%%%%%%%%%%%%%%%%%%%%%%%%%%%%%%%%%%%%%%%%%
%
\section{Introduction}
%
% Motivation
%
\IntroFirstWord
autonomous planes promise to become a ubiquitous tool in the commercial, industrial, and scientific sectors due to reduced operational costs and ever increasing robustness.
Especially the ability to map large areas and to carry out perpetual surveillance tasks, e.g. by using a  solar-powered platform, makes this type of unmanned aerial vehicles (UAVs) interesting for various applications.
While mission operation can already be completely automated \cite{Oettershagen2017a}, appropriate landing site detection (LSD) and the actual landing procedure still requires an experienced safety pilot.
Furthermore, in future fully autonomous beyond visual line-of-sight (BVLOS) operation, finding an appropriate landing spot in unstructured terrain is essential for handling emergency scenarios.

%
% Research Gap / Challenges
%
Existing LSD systems focus on the cases of  vertical takeoff and landing (VTOL) platforms, or large-scale planes, may rely on offline-computed data, or require prior knowledge about the environment.
These approaches are not suited for small-scale autonomous planes operating in unknown environments which are constrained by potentially limited energy supply and computational power.
Furthermore, their size and speed requires taking the wind into consideration, and due to their potential absence of landing gear, preferably landing in flat grass to not damage wings or fuselage.

\begin{figure}[t]
\includegraphicsHighLowRes{width=\columnwidth}{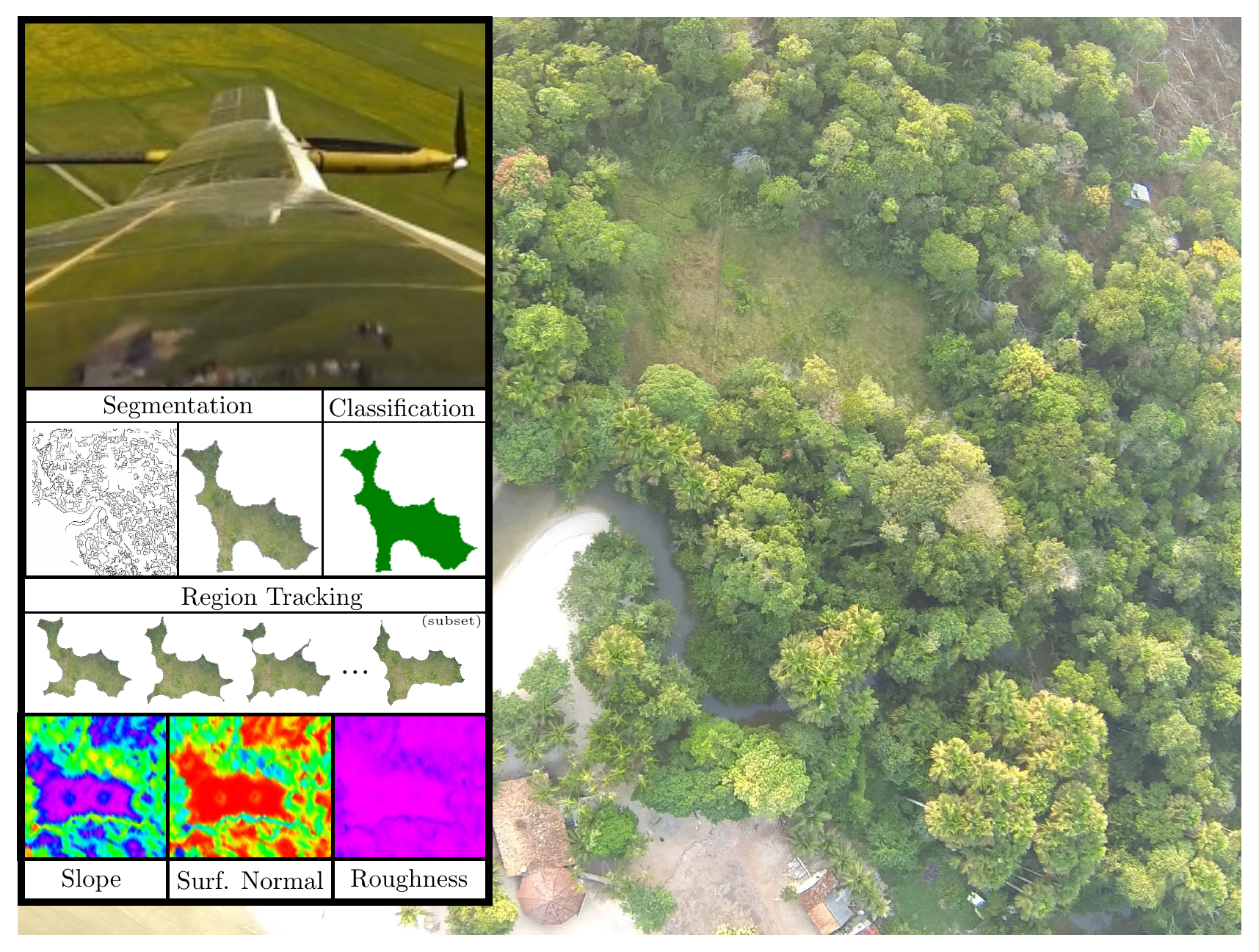}{tikz/teaser/teaser_v8}{The goal is to find the optimal landing spot while considering terrain shape, terrain texture, terrain roughness, terrain slope, surrounding obstacles, estimated local wind field, and UAV dynamics and their uncertainties.}
\label{fig:teaser}
\end{figure}

%
% Free LSD
%
The present work proposes \textit{Free LSD}, a real-time visual landing site detection and approach path computation algorithm for autonomous fixed-wing UAVs.
To keep the problem complexity manageable, potential landing sites are tracked and ranked over multiple frames.
Only the most promising landing sites are forwarded for finer-grained, $3$D processing.
No a priori data such as markers, pre-classified Digital Surface Maps (DSM), or orthomosaics are utilized which allows the framework to be operated in completely unknown terrain  as is exemplary shown in Fig.~\ref{fig:teaser}.
To the best of the authors' knowledge, this paper presents the first such system, which is also suitable for application on small-scale UAVs.
The work incorporates wind field and nearby obstacle consideration during approach path generation and decision making.
Performance of the full framework is evaluated in unknown terrain using various synthetic datasets and real-world test flights.
\section{Related Work}
Automated landing of VTOL UAVs has been considered in a broad body of works.
For instance, Desaraju et al.~\cite{Desaraju2015} propose a vision-based landing site evaluation framework to land on rooftops employing a Gaussian process to estimate the landing site confidence.
\review{Forster et al. \cite{Forster2015} present an efficient way to compute a vision-based elevation map on-board of a quadrocopter.
%equipped with a smartphone processor and using a single down-pointing camera and an IMU.
%
%The map is updated at $\unit[1]{Hz}$ using a Bayesian depth estimator based on feature tracking. This system has resulted in a robust landing of the MAV both in- and outdoors. 
%
%The work by Bosch et al. \cite{Bosch2006} does not compute a complete depth map, but instead uses a robust homography estimation on tracked dense feature points to retrieve only flat areas.
%
%These areas are marked in a stochastic grid updated at every image taken from a monocular gray-scale camera.
%
%Johnson et al. \cite{Johnson2005} tackle the problem of landing site detection for a large autonomous helicopter.
%
%A dense feature tracking algorithm followed by a structure from motion pipeline updates a terrain map from a single camera.
%
%By computing the local slope and roughness from this map, hazards such as steep slopes, rocks or cliffs can be avoided both for path planning and for landing.
%
%\igor{It's okay to cite Scherer, but strictly spoken using a previously computed point cloud kind of does not count as active landing site selection}
%Scherer et al. \cite{Scherer2010} use a previously computed dense LiDAR point cloud as an input to their pipeline.
%
%Additionally to the flatness estimate, they consider the skid contact and thus estimate the stability of the landing as well as body clearance based on a $3$D mesh estimate of the terrain for their landing site assessment.
%
Johnson et al. \cite{Johnson2002} use a LIDAR-based elevation map to compute terrain smoothness, roughness, and incidence angles to determine safe landing spots for spacecrafts.}{review1:references}
Garcia-Padro et al. \cite{Garcia-Pardo2002} introduce a contrast descriptor to land an autonomous helicopter far away from obstacles under the assumption that the terrain is flat. 
\review{Brockers \cite{Brockers2011}, Cheng \cite{Cheng2010}, and  Bosch et al. \cite{Bosch2006} make use of homography decomposition for identifying planar landing spots.}{review1:references}
Theodore et al.~\cite{Theodore2006} employ a stereo vision rig mounted on an unmanned helicopter to compute a range map and infer safe landing spots based on roughness, slope, and distance to closest obstacle.
The above approaches have in common that the main criterion for VTOL UAVs is flatness of the landing spot.
%
%By focusing on autonomous planes operating in an unknown environment, we cannot easily adapt one of these approaches but also need to consider dynamics of the plane and additional space requirements of its landing procedure.
\review{However, our application requires taking the plane dynamics and additional space requirements during landing into consideration.}{review1:references}
%

%
%%\subsection{Automated Landing in Prepared Environments}
\review{For fixed-wing platforms, most research focuses on cases where the system recognizes modified environments or man-made structures, or where the landing site is pre-defined.
Visual servoing is employed by Huh et al. to steer a small fixed-wing UAS into a red dome-shaped airbag located in an obstacle-free area \cite{Huh2010}. 
Similarly, the framework proposed by Laiacker et al. \cite{Laiacker2013} recognizes a runway from the UAV and compares it to a known model.
Given a designated, obstacle-free landing site, the height above the ground plane can be estimated using monocular visual-inertial \cite{Barber2007} or biologically inspired stereo vision \cite{Thurrowgood2014}.}{review1:references}

%For fixed-wing platforms, most research focuses on cases where the landing site is pre-defined by a GPS location, or where the system recognizes prepared environments or man-made structures such as a runway.
%%
%Huh et al. \cite{Huh2010} land a small fixed-wing UAV into a dome-shaped airbag.
%%
%This airbag is placed in the middle of an obstacle-free area such as a grass field.
%%
%Moreover, the color of the dome is red, in order for it to be easily differentiable from its surroundings.
%%
%The UAV uses visual servoing in order to determine its path towards the dome before landing by flying into the airbag.
%%
%The work of Laiacker et al. \cite{Laiacker2013} focuses on landing a large-size fixed-wing UAV.
%%
%This is achieved by recognizing a runway from the UAV by comparing it to a computer generated model of that runway in order to compute its location and orientation relatively to the UAV.
%%
%Thurrowgood et al. \cite{Thurrowgood2014} present a way to land a midsized fixed-wing UAV on a designated obstacle-free airstrip by estimating the pose of a fixed-wing UAV relatively to the ground using stereo vision algorithms.
%%
%%REMARK (IG): {Strictly spoken, the next one is not a landing site selection algorithm but merely a landing site tracking algorithm. A reviewer might potentially ask us to compare the tracking part of our pipeline with the tracking part of Barber.
%Barber et al. \cite{Barber2007} combine an optical flow sensor with an IMU in order to land a miniature fixed-wing UAV on grass \textit{given} a designated landing area.

%
% Fitzgerald's approach
%
In contrast to the aforementioned works, this paper aims at actively selecting appropriate landing spots in an \emph{unknown} environment.
This requires generation and assessment of potential candidate areas which has, to the best of our knowledge, only been discussed in two publications:
%
% REMARK (IG): Reviewers might still ask us to compare against them. I would also criticize that Fitzgerald does not consider wind effects on small UAVs, or does he?
Fitzgerald et al. \cite{Fitzgerald2005} seek to find suitable areas for crash-landing an airplane in case of emergency.
This is achieved by detecting areas without edges on a low-quality image from a defined height of $\unit[2500]{ft}$, before classifying them in order to retrieve large grass fields.
However, relying on a fixed height makes this approach disadvantageous in case of emergencies.
%
% Warrens approach
%
The closest approach to ours is presented by Warren et al.~\cite{Warren2015}. 
However, we see the following caveats that we address with the present work:
Firstly, the terrain classification is derived from stored data.
Secondly, the approach trajectory and height of nearby obstacles is only considered indirectly by the Principal Component Analysis (PCA).
Thirdly, wind is not considered which has a large effect on smaller and light-weight planes.
Finally, the approach by Warren et al. \cite{Warren2015} does not run in real-time.

\begin{figure*}[htb]
\hfill
\includegraphicsHighLowRes{width=\linewidth}{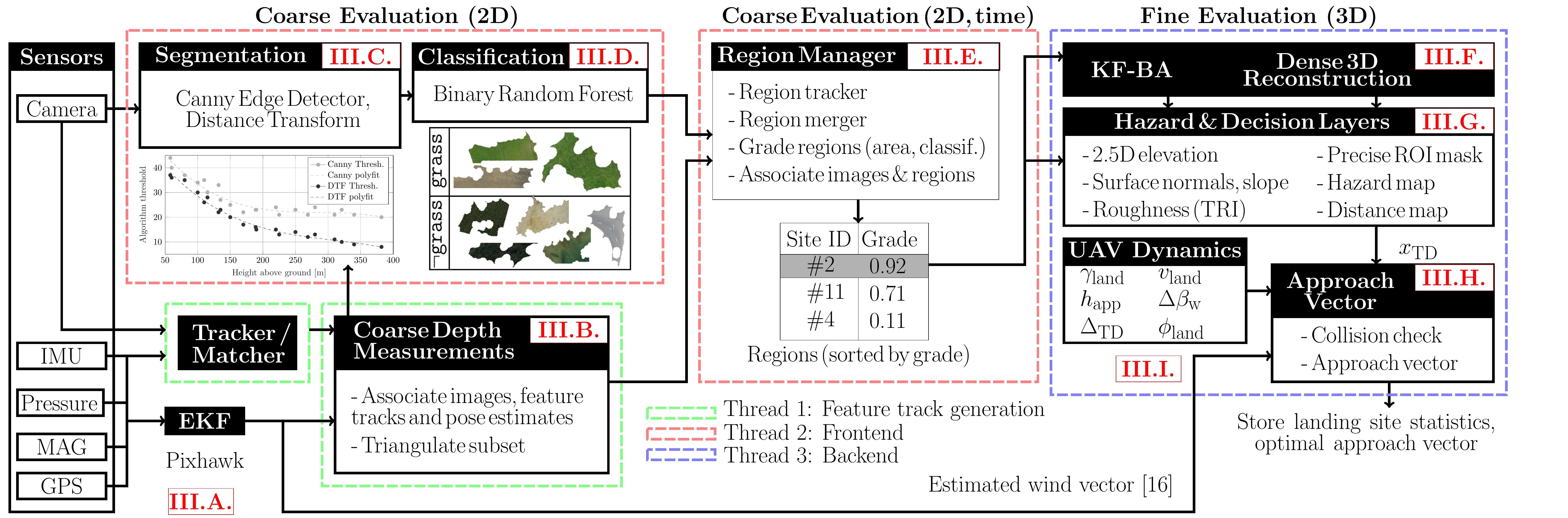}{tikz/icra18_overview_v6.pdf}{Proposed framework for prior-free landing site detection. The frontend segments, classifies, and manages the potential region of interests (ROIs). The most promising site is forwarded to the backend for finer 3D analysis and computation of approach path.}%} %\igor{Some of the text here is too small, while in some other captions it is too big. Overall, you can probably save space by unifying the size to something not much smaller than footnotesize}}
\label{fig:overview}
%
%\vspace*{\floatsep}
\vspace{-10pt}
\end{figure*}
\section{The Approach}
An overview of our proposed algorithm for the detection of landing sites is shown in Fig.~\ref{fig:overview}:
The raw image is segmented into homogeneous regions \reviews{(Sec. \ref{sec:segmentation})}{review2:fig2}{review2:section3} and classified into \verb|grass| or $\neg$\verb|grass| using a binary Random Forest (RF) classifier \reviews{(Sec. \ref{sec:classification})}{review2:fig2}{review2:section3}.
In parallel, the on-board EKF of the \textit{Pixhawk} autopilot estimates UAV pose and local wind field \reviews{(Sec. \ref{sec:state_estimation})}{review2:fig2}{review2:section3}, and depending on the provided image rate and overlap of subsequent frames, a tracker or matcher is employed to connect consecutive camera frames via feature tracks.
Resulting coarse depth measurements \reviews{(Sec. \ref{sec:coarse_depth})}{review2:fig2}{review2:section3} are used in the region manager to track region of interests (ROI) based on geometry.
The region manager \reviews{(Sec. \ref{sec:region_manager})}{review2:fig2}{review2:section3} accumulates all information about the regions and ensures consistency and uniqueness by merging regions.
Based on these metrics, a coarse grade determines which region is passed on as a candidate to the fine, $3$D evaluation backend.
This backend is periodically updated by the frontend with the $n$ most promising ROIs.
All observations of a ROI, UAV pose estimates, and previously generated feature tracks are used to perform key-frame based bundle adjustment (BA) and dense $3$D reconstruction \reviews{(Sec. \ref{sec:reconstruction})}{review2:fig2}{review2:section3}.
Metrics such as terrain slope and roughness \reviews{(Sec. \ref{sec:hazard})}{review2:fig2}{review2:section3} are derived from the 
classification results and $3$D model.
A distance-to-hazard map determines the landing spot with maximum distance to the next hazard.
Based on this touch down point, the estimated local wind field \reviews{(Sec. \ref{sec:approach_vector}, \ref{sec:wind})}{review2:fig2}{review2:section3}, and the $2.5$D elevation map, a collision-free approach path is computed.
%
%Every $N$ frames (used $N=150$) the best $n$ ROIS (used $n=1$) are enqueued for the backend. The backend then returns the ROI metrics.
%\color{black}
%
%
The final decision module outputs the landing site location, optimal approach vector, and statistics about the final landing site.
\review{The actual tracking of the final approach path is described in \cite{Oettershagen2017a}.}{review1:landing}
The metrics of the best landing sites are stored to be able to land quickly in the case of an emergency.
\subsection{State Estimation}\label{sec:state_estimation}
The state estimator on the \textit{Pixhawk} autopilot estimates body poses, velocities, IMU biases, and the wind field using GNSS, IMU, magnetometer, and pressure measurements~\cite{Leutenegger2014}.
The camera pose estimates are forwarded to the on-board computer which associates camera poses to the corresponding images based on the pre-calibrated camera-IMU transformation \cite{Furgale2013}.
These camera pose estimates are used as priors in the bundle adjustment if an area was marked as potential landing spot.
Additionally, feature tracks are generated using, depending on the provided framerate, a Kanade-Lucas-Tomasi (KLT) \cite{Lucas1981} feature tracker or matcher.
These feature tracks are used to generate coarse depth measurements (cf. Sec. \ref{sec:coarse_depth}) for region tracking in unknown terrain and in the bundle adjustment of the backend thread (cf. Sec. \ref{sec:reconstruction}).

\subsection{Coarse Depth Measurements}\label{sec:coarse_depth}
To obtain a segmentation that is robust to height changes as well as for geometric region tracking, coarse depth measurements are required in the frontend (cf. Sec. \ref{sec:region_manager}).
Since our system is designed to operate in unknown terrain without a priori data, the depth measurements need to be retrieved at runtime\footnote{Depending on the application and flight altitude, the coarse depth measurements could alternatively be obtained from a ground plane approximation.}.
One possibility would be to triangulate a few features at every step and build up a mesh by using, for instance, Delaunay triangulation~\cite{Weiss2011}.
However, to obtain depth measurements at a given pixel, computationally expensive ray-casting queries would be required.
Furthermore, a depth image obtained from two views from a virtual stereo rig based on unoptimized camera pose estimates is prone to errors since we assume a noisy low-level state estimator.
\review{Instead, we take advantage of the feature tracking thread that is running in parallel: To map from $2$D to $3$D coordinates, the $N$ feature tracks closest to the queried keypoint location are determined.
The final height of the requested keypoint location is obtained by performing multi-view triangulation of the $N$ nearby tracks and inversely weighting the resulting triangulated landmark heights by their distance to this keypoint.}{review2:depth_measurements}
%
%Instead, we take advantage of the feature tracking thread that is running in parallel, and if demanded, query the closest $N$ feature tracks in image space and perform multi-view triangulation and weigh the elevation by the inverse distance of the nearby tracks.
%
\subsection{Region Segmentation} \label{sec:segmentation}
\begin{figure}[htb]
\begin{subfigure}{0.22\columnwidth}
\includegraphicsHighLowRes{width=\linewidth}{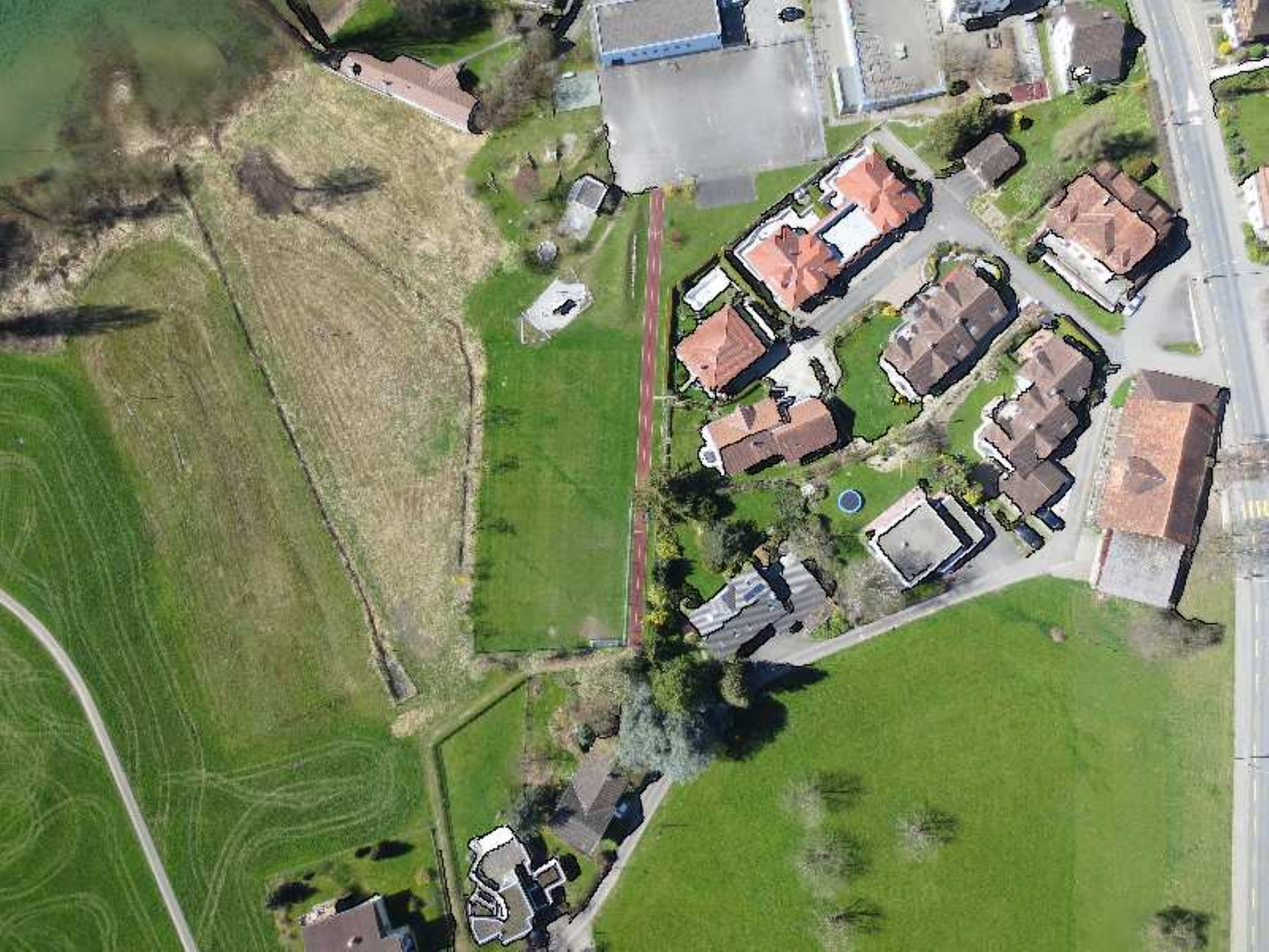}{images/high-res/1_compressed.pdf}{}
\label{fig:seg_1}
\end{subfigure}
\hfill
\begin{subfigure}{0.22\columnwidth}
\centering
\includegraphicsHighLowRes{width=\linewidth}{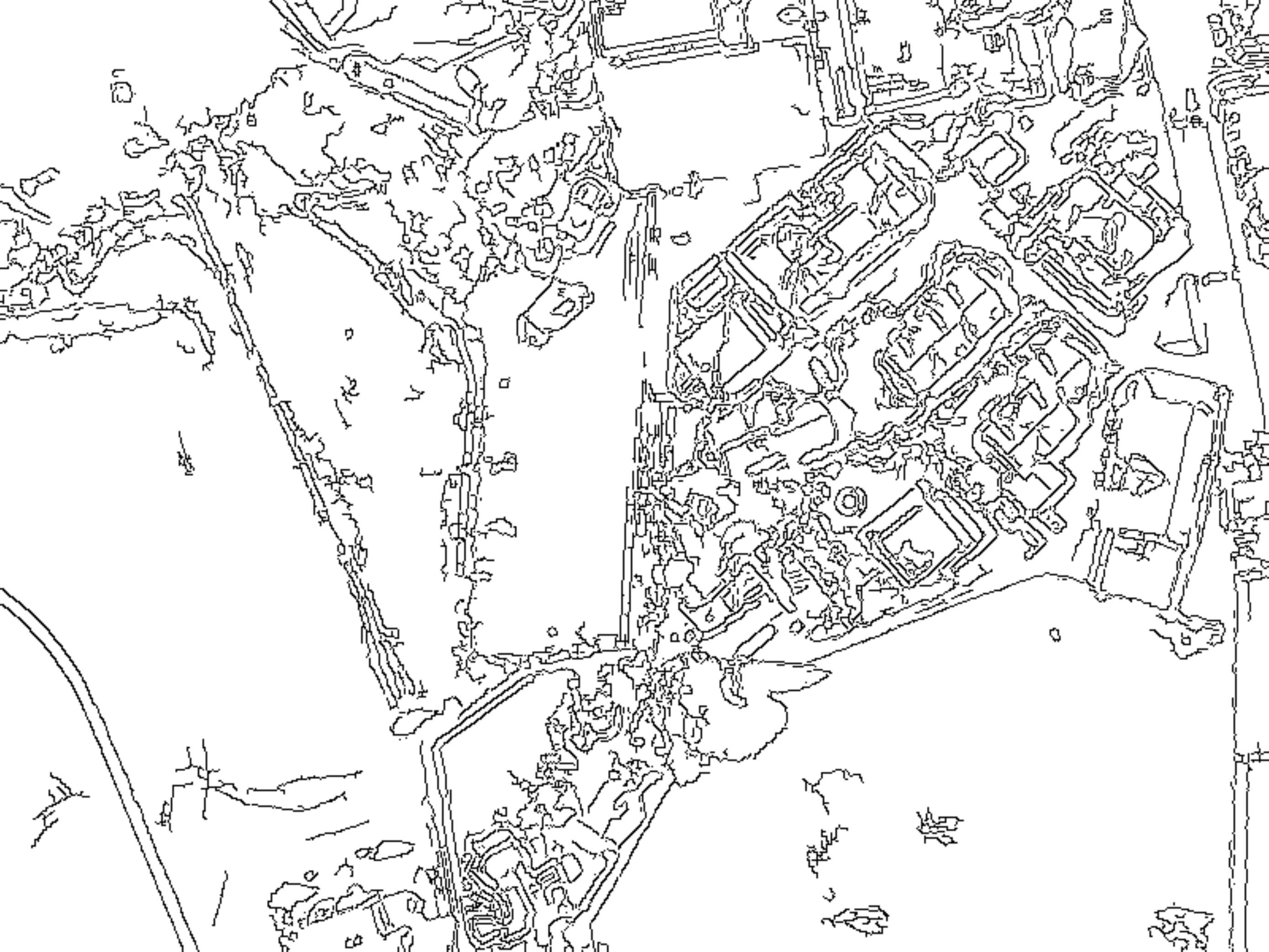}{images/high-res/3.pdf}{}
\label{fig:seg_2}
\end{subfigure} 
\hfill
\begin{subfigure}{0.22\columnwidth}
\centering
\includegraphicsHighLowRes{width=\linewidth}{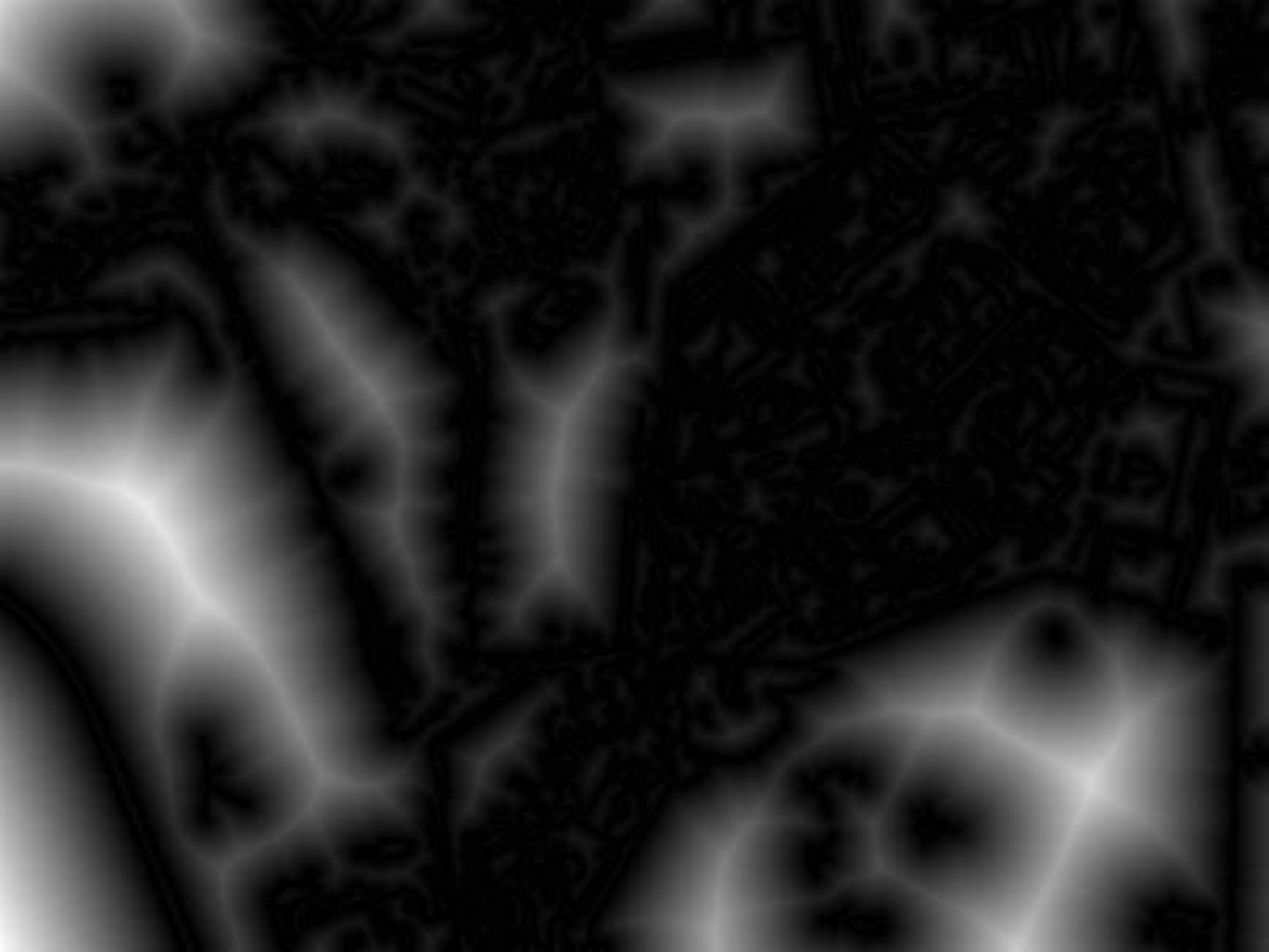}{images/high-res/4_compressed.pdf}{}
\label{fig:seg_3}
\end{subfigure}
\hfill
\begin{subfigure}{0.22\columnwidth}
\centering
\includegraphicsHighLowRes{width=\linewidth}{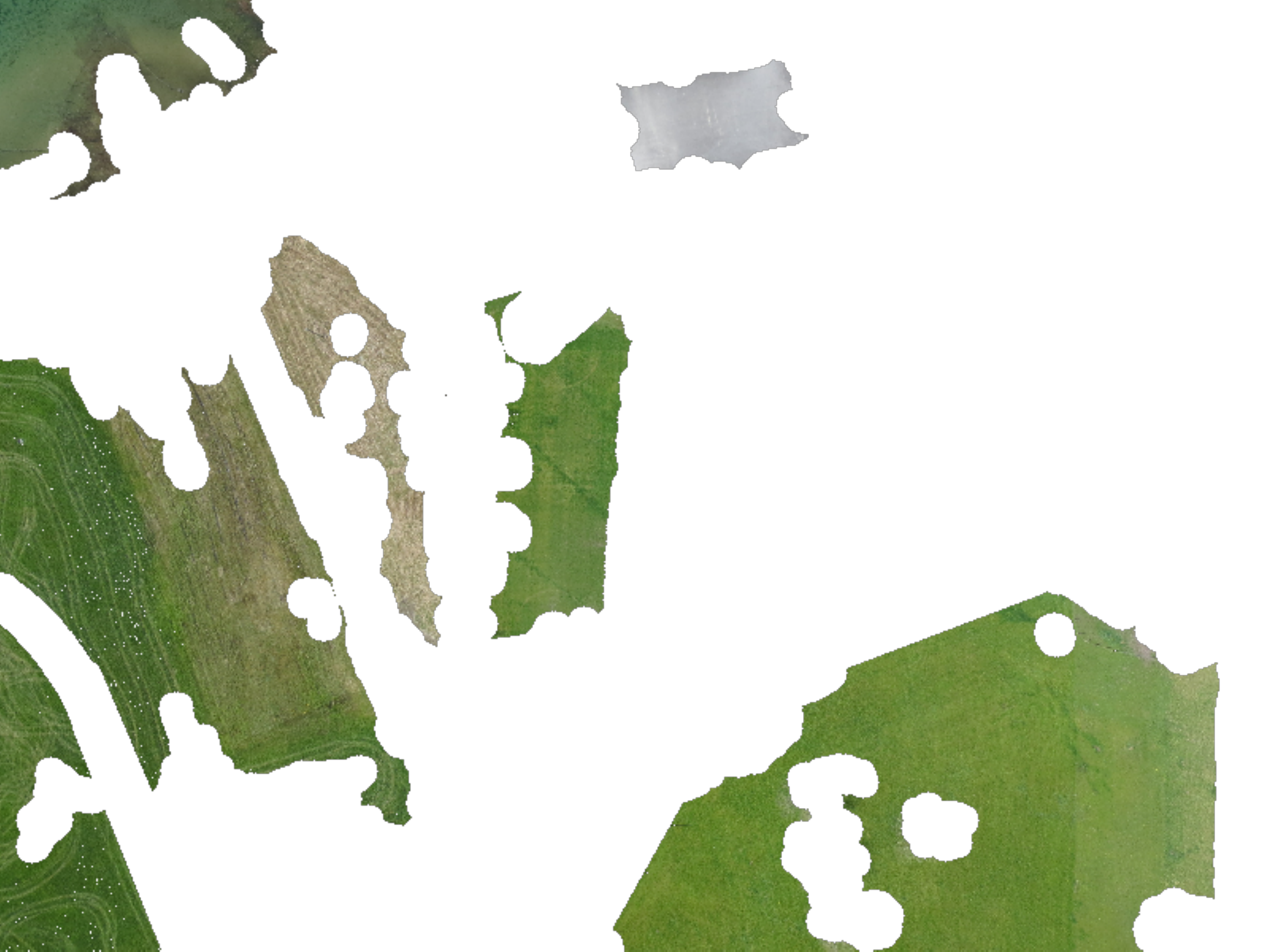}{images/high-res/5.pdf}{}
\label{fig:seg_4}
\end{subfigure}
\hfill
%
%\begin{figure}[htb]
%\begin{subfigure}{0.48\columnwidth}
%\includegraphics[width=\linewidth]{images/1_merlischachen_original.png} 
%\caption{}
%\label{fig:seg_1}
%\end{subfigure}
%%
%\hfill
%%
%\begin{subfigure}{0.48\columnwidth}
%\centering
%\includegraphics[width=\linewidth]{images/3_distTransform_in.png} 
%\caption{}
%\label{fig:seg_2}
%\end{subfigure} \\
%%
%\begin{subfigure}{0.48\columnwidth}
%\centering
%\includegraphics[width=\linewidth]{images/4_distTransform_out.png} 
%\caption{}
%\label{fig:seg_3}
%\end{subfigure}
%%
%\begin{subfigure}{0.48\columnwidth}
%\centering
%\includegraphics[width=\linewidth]{images/5_distTransform_segmented_transparent.png} 
%\caption{}
%\label{fig:seg_4}
%\end{subfigure}

%\hfill
%%
%\begin{subfigure}{0.24\columnwidth}
%\centering
%\includegraphics[width=\linewidth]{images/5_distTransform_segmented.png} 
%\caption{}
%\label{fig:seg_4}
%\end{subfigure}
%
\caption{Region segmentation: (a) Original input image, (b) Canny edges, (c) Distance transformation, (d) Segmented regions. Regions with a small area are rejected already at this step.}
%\vspace{-10pt}
\end{figure}

The Canny edge detector \cite{Canny1983} is applied to the grayscale spectrum of the raw image (cf. Fig.~\ref{fig:seg_1}).
The result is shown in Fig.~\ref{fig:seg_2}.
Next, the distance transform \cite{Felzenszwalb2012} is applied to compute for every pixel the distance to the closest non-zero pixel or Canny edge.
The distance map, as shown in Fig.~\ref{fig:seg_3}, is then thresholded to obtain homogeneous regions (cf. Fig. \ref{fig:seg_4}).
Note that high contrast obstacles, such as the trees \review{in the lower right section of the images}{review2:resolution}, are often already identified at this early stage.
The threshold in the Canny edge detector and the distance transform is computed from a function of height, to ensure that the same areas are segmented independently of the UAV's altitude above ground\footnote{The height-dependent thresholds were approximated by $p_\text{canny}(h)=-1.72e-06h^3+0.00148 h^2 - 0.43 h+62.97$, and $p_\text{dtf}(h)=-1.23e-06 h^3 + 0.0011 h^2-0.39h + 56.82$ as shown in Fig.~\ref{fig:overview}.}.
\review{The thresholds are derived from Google Earth imagery and span an altitude range of $58$-$\unit[382]{m}$ above ground.
For reference, the nominal flight altitude of the deployed UAVs in this publication is between $50$ and $\unit[250]{m}$}{review1:thresholds}.
\newlength\figureheight
\newlength\figurewidth
\setlength\figureheight{4.5cm}
\setlength\figurewidth{0.85\linewidth}
%\begin{figure}[htp]
%\input{matlab/canny_distanceTF_height_dependency/canny_dtf_height_dep.tikz}
%\caption{Height dependency of Canny edge detector and distance transform}
%\label{fig:canny-dtf}
%\end{figure}
%
%
%
%\begin{figure}[htb]
%%
%\includegraphics[width=1\linewidth]{tikz/training_sample_patches.pdf}
%%
%\caption{Region patches used to train the binary classifiers. The patches are extracted from various datasets (different cameras).}
%%
%\label{fig:region_patches}
%%
%\end{figure}
%
\subsection{Region Classification}\label{sec:classification}
The segmentation module presented in the previous section only ensures that the extracted area is homogeneous.
In the classification step, the texture and color properties of the homogeneous area are extracted to classify the regions into \verb|grass| or $\neg$\verb|grass| as illustrated in Fig. \ref{fig:overview}.
For this purpose, we employ a binary Random Forest  (RF) \cite{Breiman2001} classifier which takes the segment from Fig.~\ref{fig:seg_4} and predicts the binary label.
The classifier is trained based on a set of features extracted from the homogeneous regions.
The parameters of the classifier, that is, the maximal tree depth and the number of samples needed per branch, are optimized on the training data by $10$-fold cross-validation.
The ground truth for the classification is established as follows: Homogeneous regions are obtained by the described segmentation algorithm. After visual inspection, the region is manually labeled as \verb|grass| or $\neg$\verb|grass|.
\begin{figure}[htb]
\includegraphicsHighLowRes{width=\linewidth}{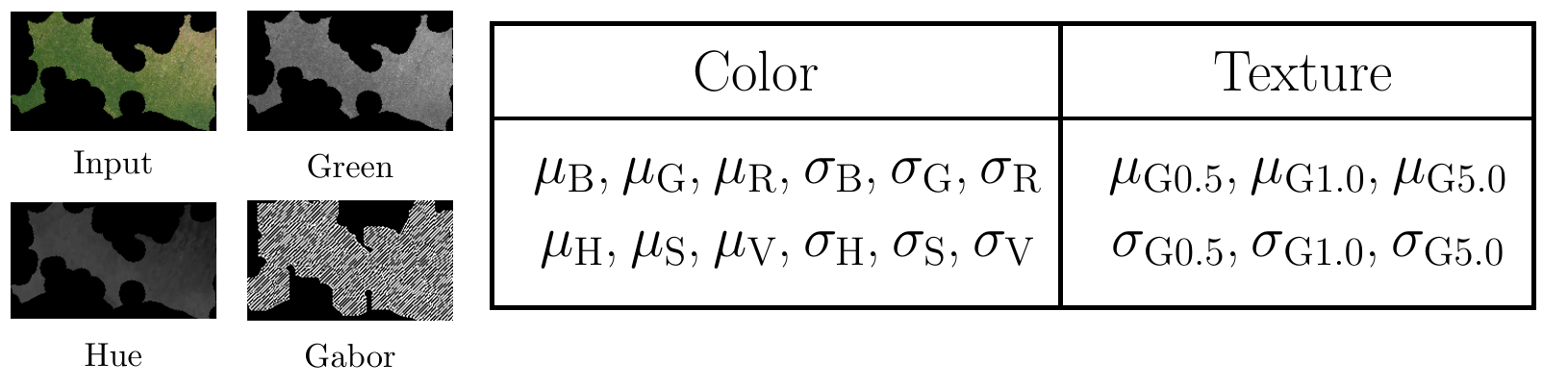}
{tikz/features_classification_3.pdf}{Features used for binary classification of homogeneous regions. Note that the Gabor feature applied in form of a convolutional filter expands the area, but only the information within the ROI mask is used to compute mean and standard deviation.}
\label{fig:features}
\end{figure}
\subsubsection{Feature Space}
For each segmented ROI, twelve color and six texture features, are extracted as summarized in Fig.~\ref{fig:features}.
\paragraph{Color}
For each sub-image, the mean and standard deviation for all three color channels are computed across the complete segmented ROI.
This is performed not only in the standard RGB color space, but also in the HSV space.
In many computer vision applications, the HSV space has proven to be less sensitive to lighting conditions, when comparing to RGB \cite{Agoston2005}.
While the classifier performs better using the HSV color space than
RGB only, it performs even slightly better when using the features extracted from both: The classification error for only using RGB is $\unit[15.36]{\%}$, $\unit[14.26]{\%}$ for HSV, and $\unit[14.12]{\%}$ for RGB and HSV.
We hence get a total of (2 color spaces) $\times$ (3 colors) $\times$ (2 features per color) = 12 color features which are computed for every sub-image.
While more advanced color features can be extracted, e.g. using various combinations of color histograms \cite{Smith2002}, we here only rely on these very simple features for low computational costs.
\paragraph{Texture}
Color features are sensitive to illumination and viewing angle.
To better assess the spatial arrangement of intensities in an image patch we additionally compute texture descriptors. 
%We therefore combine these features with texture features.
%
%Texture descriptors are a way of characterizing the spatial arrangement of intensities in an image.
%
%The combination of color and texture descriptors offers a more complete
%description of how humans perceive and discriminate different real-world textures such as grass, water or trees.
%
%In order to quickly extract and characterize texture from an extracted subimage, we use Gabor filters.
%
For this task, we employ Gabor filters~\cite{Gabor1946}, linear filters related to the Gabor Wavelet that extract texture features from gray-scale imagery~\cite{Jain1990} more efficiently than alternatives, such as Local Binary Patterns (LBP)~\cite{Ghiasi2013, Pietikaeinen2002}.
%These are a class of linear filters related to the Gabor Wavelet introduced by Gaborand which can be used, amongst others, for edge detection.
%
%While more recent texture descriptors such as local binary patterns (LBP) have proven to be performing well for aerial image classification problems~\cite{Ghiasi2013, Pietikaeinen2002}, 
%
The following parameters for phase offset $\varphi$, standard deviation of the Gaussian function $\sigma$, and spatial aspect ratio $\gamma$ are used: $\varphi = 0$, $\sigma = 4$ and $\gamma = 0.02$.
%
%\begin{table}
%\footnotesize
%\begin{tabular}{|l|c|r|l|c|r|r|} \hline
%  & BGR & $\neg$B$\neg$G$\neg$R & H,S,V& $\neg$H,$\neg$S,$\neg$V & $\text{G}^1$ & $\text{G}^1$\\
%med.  &0.51,0.61,0.63 & & & & &\\  
%1st  & & & & & & \\  
%3rd   & & & & & &\\  
%   \hline
%\end{tabular}
%\end{table}
%%
%\setlength\figureheight{1.5cm}
%%
%\begin{figure}[htb]
%%
%\begin{subfigure}{\linewidth}
%%
%\setlength\figurewidth{0.85\linewidth}
%\input{matlab/boxplot_feature_bgr_v2.tikz} \\
%\setlength\figurewidth{0.22\linewidth}
%\input{matlab/boxplot_feature_h_v2.tikz} 
%\setlength\figurewidth{0.42\linewidth}
%\input{matlab/boxplot_feature_sv_v2.tikz} \\ 
%\setlength\figurewidth{0.85\linewidth}
%\input{matlab/boxplot_feature_gabor_v2.tikz} 
%%
%\end{subfigure}
%%
%\caption{Color and texture mean features used to classify the regions in ``non-grass" ($\neg g$) and ``grass" ($g$). The features based on the standard deviation are not shown. }
%%
%\label{fig:feature_boxplot}
%%
%\end{figure}
%
The orientation $\theta$ in which the edges are detected is not important in our case, since we try to detect rotation-independent descriptors.
The Gabor filtered images are computed by applying a convolutional filter in four directions $\theta \in \{0, \pi/4, \pi/2, 3\pi/4\}$ and taking the mean of the extracted values.
This approach yields three Gabor filtered images for the wavelengths $\lambda \in \{0.5, 1, 5\}$.
The final descriptors used in the classifier correspond to the mean and standard deviation of each of these Gabor filtered images, hence a total of six texture descriptors.
%
%For reference, in Section~\ref{ssec:exp-cla} the Random Forest classifier is compared to a three-layered neural network which is based on a multilayer perceptron model \color{red}(\emph{opencv} \cite{Itseez2015} impl.)\color{black}.
%
%The color and texture mean features used to classify the regions in $\neg$\verb|grass| or \verb|grass|, computed on a training set, are shown in Fig.~\ref{fig:feature_boxplot}.
%
\subsection{Region Manager: Tracking, Merging and Updating}\label{sec:region_manager}
%
%
%\begin{figure}[htb]
%\begin{minipage}{0.49\textwidth} 
%\includegraphics[width=\linewidth]{tikz/images/roi_definition.png} 
%\label{fig:seg_1}
%\end{minipage}
%\hfill
%\begin{minipage}{0.49\textwidth} 
%\includegraphics[width=\linewidth]{tikz/ROI_tracking_2.pdf} 
%\label{fig:seg_1}
%\end{minipage}
%\end{figure}
\subsubsection{Tracking and Updating of ROIs}
\begin{figure}
    \centering
    \begin{minipage}[b]{.45\linewidth}
        \centering
    \centering\includegraphics[width=1\linewidth]{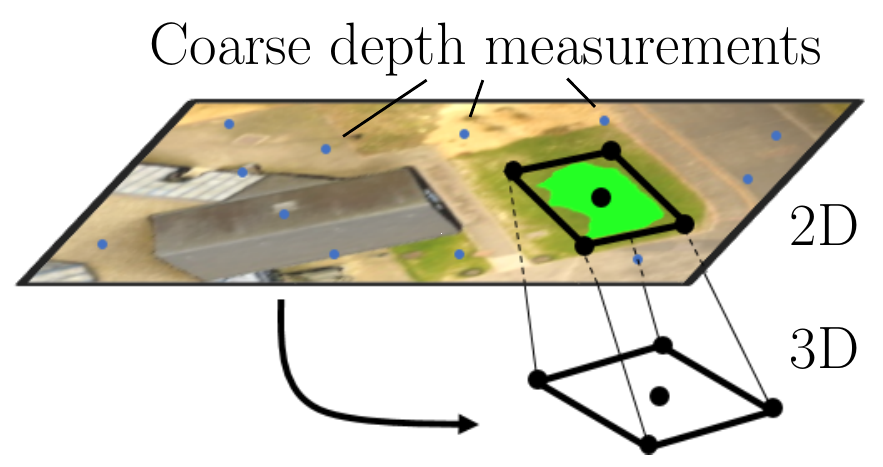}{}    
    \end{minipage}%
    \hfill%
    \begin{minipage}[b]{.45\linewidth}
    \centering
        \centering
\centering\includegraphics[width=1\linewidth]{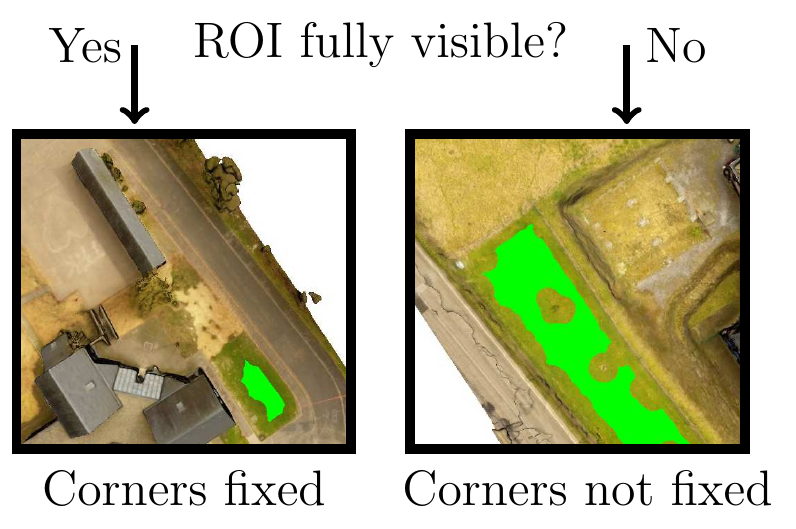}{}
%\label{fig:roi_tracking}
    \end{minipage}
    \begin{minipage}[t]{.45\linewidth}
        \caption{ROI initialization}
        \label{fig:roi_init}
    \end{minipage}%
    \hfill%
    \begin{minipage}[t]{.45\linewidth}
        \caption{ROI tracking}
        \label{fig:roi_tracking}
    \end{minipage}%
    \vspace{-8pt}
\end{figure}

%\begin{itemize} 
%\item Fig. 1: fine classification mask, consisting of a set of $2$D points, to minimum-area enclosing rectangle using rotating caliper \cite{Toussaint1983,Itseez2015}. Simplifies tracking and increase robustness with respect to segmentation output or height approximation.
%\item Fig. 1: rectangle (2D) to (3D) using coarse depth measurements.
%\item Fig,2: ROI Init: ideally Case(fully visible) all corners of rectangle in current image, fix corners. Center of ROI+ inside corners of ROI (winding) -> Update: nclass, nobs, grade); else init new ROI 
%\item ROI Init: Case(not fully visible): one or more corners outside of current image, corners are set but not fixed. ROI+ associated to ROI with non-fixed corners -> Update: Rectangle that incorporates all 8 points \cite{Toussaint1983,Itseez2015}
%\end{itemize}
%%
As illustrated in Fig.~\ref{fig:roi_init}, the classification and segmentation module forwards the contours of a fine classification mask, defined by a set of $2$D points, to the region manager.
To simplify tracking and to increase the robustness with respect to impairing factors\footnote{E.g. the depth approximation introduced by the coarse depth measurements utilized for the $2$D to $3$D projection.}, the fine classification mask is approximated by the minimum-area enclosing rectangle using the rotating caliper method \cite{Toussaint1983,Itseez2015}.
Next, the $2$D positions of the four corners and centroid of the rectangle are projected into $3$D based on the available coarse depth measurements (cf. Sec. \ref{sec:coarse_depth}).
As depicted in Fig.~\ref{fig:roi_tracking}, two cases are distinguished for initializing and updating ROIs: In the first case, the ROI is fully visible, i.e. all $2$D corners are within the current image. If so, the corresponding $3$D corners are fixed and ROI statistics ($n_\text{grass}$, $n_\text{obs}$, grade, cf. Sec. \ref{sec:grade}) are set.
In a subsequent frame, a re-detection is triggered if the centroid of the ROI in the current frame is within the corners of an existing ROI, which is determined by the winding number method~\cite{Weiler1994}.
In this event, only the tracked ROI statistics are updated.
In a second case, if the ROI is not fully visible, i.e. one or more corners are on the border of the image, the $3$D corners are set but not fixed.
If, in a subsequent re-detection, the ROI is again not fully visible the corners are updated by the vertices of the rectangle that incorporates all $8$ corners \cite{Toussaint1983,Itseez2015} until the ROI is fully visible and the first case applies.
%%For each segmented area in the image the UTM coordinates of the four corners of the rotated rectangle which embodies the ROI are saved, if all of them are inside the raw image. \red{UTM? are the corners estimated by stereo and then referenced in the current estimated pose in UTM coordinates?}\timo{No, from rough altitude estimate (take-off alt. as reference). I'm investigating if it's necessary to make it robuster via tracking/triangul. (We only have a mono setup.)}
%%
%Whenever a new area is detected, we compare the projected coordinates of the center of the ROI with the corners of all already tracked areas.
%%
%
\subsubsection{Merging of tracked ROIs}
%
%\begin{figure*}[htb]
%%
%\centering\includegraphics[width=\linewidth]{tikz/decision_layers.pdf}\par
%%
%\caption{Hazard (or Decision) Layers. \timo{Will be filled with output from framework}}
%%
%%\vspace*{\floatsep}
%\label{fig:decision_module}
%%
%\end{figure*}
%
It can occur that two tracked regions of interest correspond in fact to the same landing area.
An example for this would be if only half of the area is detected in a series of subsequent frames, while the other half is detected later on in other frames.
%
%The initial position of the area centers and corners would hence be different, and the pipeline could not determine that those are in fact belonging to the same area.
%
However, in following images, the UAV could detect the complete area.
Without any merging, the pipeline would update values such as the corner position or the area size for one of these two areas because the projection of the newly detected center point is placed inside it, while the other half would remain unchanged.
%
%This yields to one tracked area being inside the other one, a redundancy which we want to avoid.
%
To avoid this duplication, every time the corner positions of a tracked ROI get updated, we verify for each ROI in the tracker if it belongs to that area, i.e. if the center of the newly updated area is in-between the four corners of the tracked ROI.
If that is the case, the two ROIs are merged: the corner positions are set to the ones of the largest area and the grade is updated accordingly.
%

%It can happen that nearby, but dissimilar, regions meet the geometric condition for region merging due to erroneous pose estimates or a low altitude-to-terrain-elevation ratio.
%%
%For improved robustness, the feature vector of the region to be merged and the first region observation, which serves as reference, are compared.
%%
%In particular, for each region, the mean is set by the feature vector extracted from the first region observation, the standard deviation is obtained from cross-validation over training patches.
%\timo{Consider using multivariate Gaussian anomaly rejection. But maybe overkill.}
%%
\subsubsection{Grading of tracked ROIs}\label{sec:grade}
The tracked ROIs are ranked according to a cost function assigning a grade to each landing spot.
%
% (This can be seen easily from the cost function)
%A higher grade means the area is more suited to land on.
%
The grading function makes use of metrics computed for each tracked region: The area $\text{A}$ spanned by the four projected corners, the number of images in which the ROI has been classified as grass $n_\text{grass}$, and the total number of images in which it has been observed $n_\text{obs}$.
The grade is zero if $\text{A} < \text{A}_\text{min} \text{ or } n_\text{obs} < n_\text{obs,min}$ and $ n_\text{grass} n_\text{obs}^{-1}$ otherwise. 
%Our coarse grading function is given by:
%
%\begin{align}
%    f(x)= 
%\begin{dcases}
%    0,& \text{if A} < \text{A}_\text{min} \text{ or } n_\text{obs} < n_\text{obs,min} \\
%    \frac{n_\text{grass}}{n_\text{obs}},              & \text{otherwise}
%\end{dcases}
%\end{align}
%
In order to reduce the computational load, only the $20$ ROIs with the highest grade are retained.
This is implemented in form of a FIFO buffer in order to first remove regions which have not been detected or recognized recently.
\subsection{Dense 3D Reconstruction}\label{sec:reconstruction}
\review{To reduce the computational burden, the subset of frames is iteratively selected for pose refinement and dense reconstruction as follows:
The first pose is set as key-frame (KF). Then the next frame for which the feature track connection count first drops below $30$ is determined.
The predecessor to this frame is the next KF if the baseline is larger than a minimal baseline.
Next, for every KF, the best suited stereo-pair is selected based on baseline, epipolar and viewing cost \cite{Jaeger2015}.
The selected set of poses is refined by incorporating pre-computed pose priors and feature tracks (cf. Section \ref{sec:state_estimation}).
Finally, the optimized poses are used for planar rectification \cite{Fusiello2000, Jaeger2015} in combination with Semi-Global Block-Matching (SGBM \cite{Itseez2015}).
As described in Section \ref{sec:hazard}, inverse distance weighting (IDW) is used to convert from $3$D point cloud to $2.5$D elevation map  which  smoothes the depth estimates.}{review2:mdr}

%As input, this module receives all images associated to a homogeneous ROI, the camera pose estimates corresponding to each image, and the feature tracks connecting the subsequent camera frames.
%
%Based on the camera pose priors and feature tracks, this module initializes landmarks by triangulation and performs bundle adjustment .
%
%Whenever a region, that previously fell out of the field of view of the camera, is redetected, we merge tracks and landmark observations based on descriptor matching and optimize using batch keyframe based \color{red} bundle adjustment \footnote{Fixed camera intrinsics and camera-IMU extrinsics.} (\textit{ceres} \cite{Agarwal}, maplab\color{black}). \color{red} \color{black}
%
%
%\color{red}
%Keyframe selection strategy. Stereo vision frame selection strategy: 
%\color{black}
%
\subsection{Hazard and Decision Layers}\label{sec:hazard}
\begin{figure}[htb]
\centering\includegraphicsHighLowRes{width=1\linewidth}{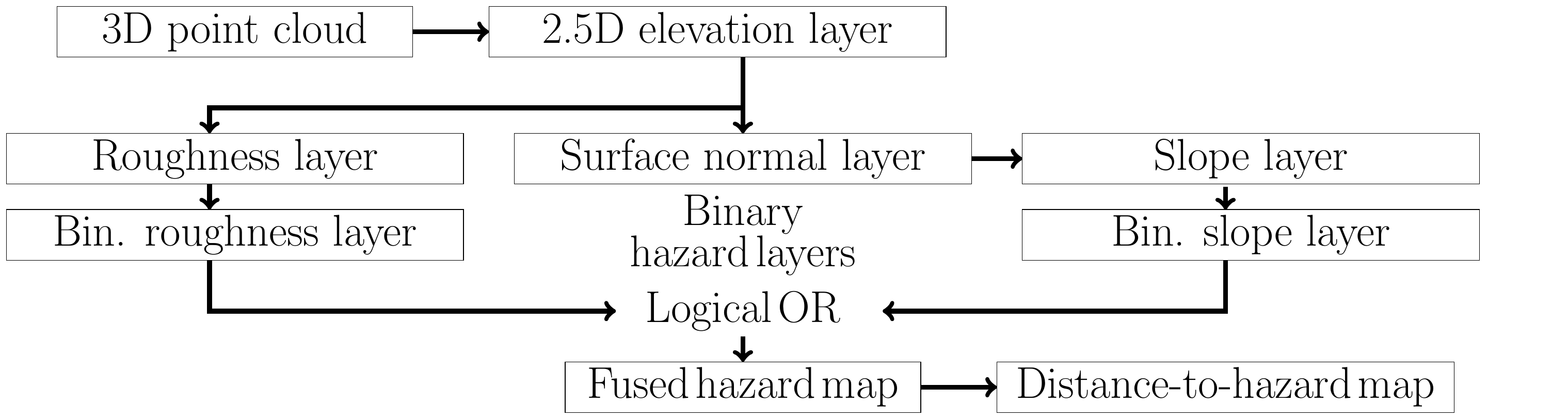}{tikz/decision_layers_v5.pdf}{Hazard and Decision Layers.}
%
%\vspace*{\floatsep}
\label{fig:decision_module}
\end{figure}

This module uses the dense point cloud as input in order to evaluate the landing spot with respect to potential hazards such as terrain slope and terrain roughness.
The data flow is presented in Fig.~\ref{fig:decision_module}, \review{for a sample visualization we refer to Fig.~\ref{fig:synth}.}{review2:Fig9}
To avoid the high memory load introduced by calculations involving the dense $3$D point cloud, we convert to a $2.5$D grid-based elevation map~\cite{Fankhauser2016} \review{(Fig.~\ref{fig:synth_c}).}{review2:Fig9}
The elevation of each cell is computed using KD-tree based \cite{Blanco2014} IDW in a radius around the cell.
From the $2.5$D elevation layer, the surface normal in $z$-direction $n_z$ of cell $c_{ij}$ is computed based on the current cell and the 8-nearest neighbor cells using PCA.
From the surface normal layer, the cell's slope $\alpha_{ij}$ with respect to the ground plane is obtained from $ \alpha_{ij} = \arccos(n_{z,ij})$ \review{(Fig.~\ref{fig:synth_d}).}{review2:Fig9}
Terrain roughness is identified as a second hazard.
The terrain ruggedness index (TRI)~\cite{Warren2015} is computed based on the elevation difference to the $8$ adjacent cells and allows, for instance, to differentiate between flat grass, crops, or forest regions \review{(Fig.~\ref{fig:synth_e}).}{review2:Fig9}
A fine classification mask is computed as logical OR operation of all fine grass classification masks associated with this ROI.
All hazard layers are only evaluated in the cells that have been classified as grass in at least one observation.
Next, the hazard layers are transformed into binary layers using thresholds that are acceptable for the UAV \review{(Fig.~\ref{fig:synth_f}).}{review2:Fig9}
The binary hazard layers are then fused using the logical OR operation.
In order to find safe and contiguous landing paths, we then apply the distance transform \review{(Fig.~\ref{fig:synth_g})}{review2:Fig9} to the fused binary hazard map.
This yields, for every cell, the distance to the closest hazard.
Further decision layers, such as a probabilistic point cloud or classification uncertainty layer, could easily be incorporated.
%
%Note that more decision layers, such as point cloud uncertainty or a finer texture classification layer, could easily be added.
%
\subsection{Landing Approach Vector}\label{sec:approach_vector}
The question remains from which direction the landing spot is to be approached while circumventing the surrounding hazard(s).
\begin{figure}[htb]
\includegraphicsHighLowRes{width=1\linewidth}{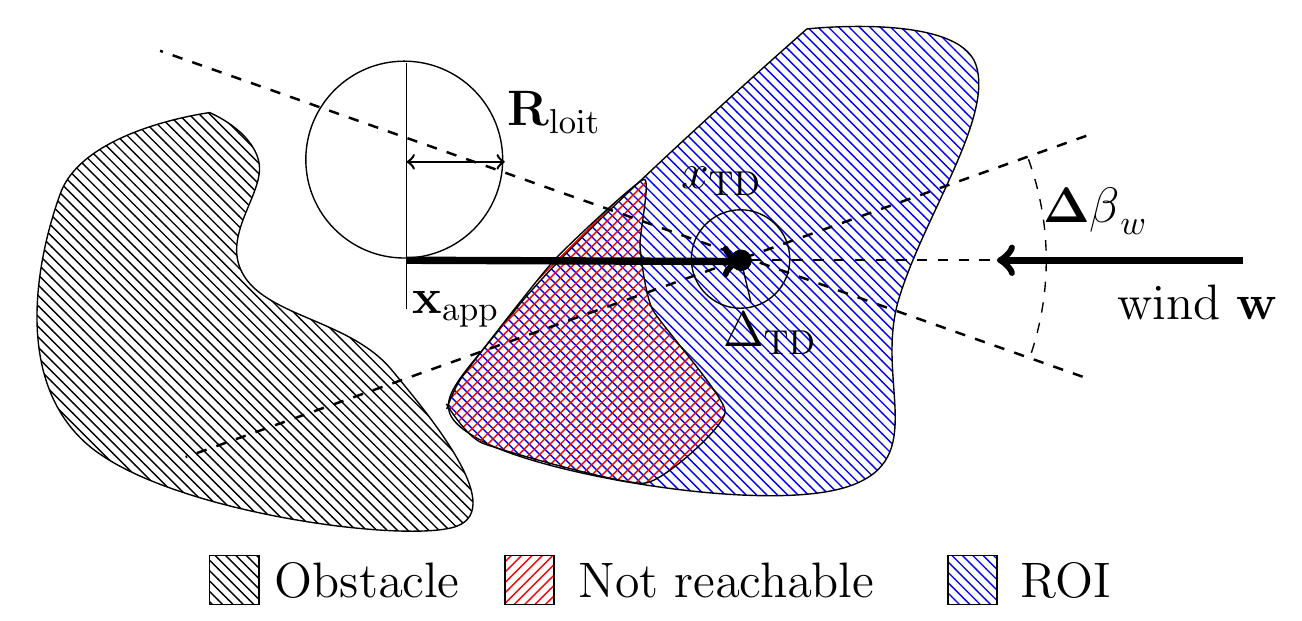}{tikz/approach_vector_v5.pdf}{Computation of the landing approach vector while considering the local wind field as well as hazards surrounding and within the landing region (ROI). }
%\igor{This graphic seems to break the page layout when built on my system (which surprisingly only happens if this ugly comment block is removed)}}
\label{fig:approach_vector}
\end{figure}
The local wind field, which is estimated in real-time by the on-board EKF, constrains the approach vector as illustrated in Fig. \ref{fig:approach_vector}.
Small-size fixed wing UAVs need to land against the wind direction in order to minimize the distance required for landing and to remain in a safe ground velocity region.
Furthermore, we consider nearby obstacles obscuring the landing field based on the maximum descent rate of the UAV as well as obstacles in the landing region which are encoded in the distance map.
Based on these considerations, the landing approach path is computed \cite{Oettershagen2017a}:
\begin{align}
x_\text{app} &= \frac{v_\text{land} \cos(\gamma_\text{land}) - w \cos(\Delta \beta_{w})}{v_\text{land}\sin(\gamma_\text{land})} h_\text{app} \nonumber\\
R_\text{loit} &= \frac{(v_\text{land}\cos(\gamma_\text{land})+w)^2}{g\tan(\phi_\text{land})}  \label{eq:approach_path}
%\Delta &= R_\text{loit} + x_\text{app} \label{eq:approach_path}
\end{align}
with
\scriptsize
\begin{align*}
x_\text{TD}&: \text{touch down point} & \Delta_\text{TD}&: \text{touch down uncertainty ($\unit[10]{m}$)}\\
w&: \text{wind magnitude (estimated)} &
v_\text{land}&: \text{airspeed ref. ($\unit[13]{m/s}$)}  \\
\gamma_\text{land}&: \text{flight path angle ref. ($\unit[4]{\deg}$)} &
\Delta\beta_w&: \text{crosswind uncertainty ($\unit[30]{\deg}$)} \\
h_\text{app}&: \text{altitude approach ($\unit[12]{m}$)}
& \phi_\text{land}&: \text{maximum bank angle ref. ($\unit[11]{deg}$),}
\end{align*}
\normalsize
\review{and approach vector $\mathbf{x}_\text{app} = x_\text{TD} + \begin{bmatrix} x_\text{app}\tilde{w}_x,\text{ } x_\text{app}\tilde{w}_y, \text{ }h_\text{app} \end{bmatrix}^\top{}$, where $\tilde{\mathbf{w}}=\begin{bmatrix} \tilde{w}_x,\text{ } \tilde{w}_y \end{bmatrix}^\top{}$ is the normalized $2$D wind vector.}{review1:approach_path}
%
%The listed numeric values are used for the research UAV Techpod as described in Section \ref{rw_experiments}. 
The numbers in parentheses are sample values used for the research UAV \emph{Techpod}.
Based on the distance-to-hazard map we efficiently take the touch down uncertainty into consideration.
In particular, starting from the safest touch down point, we check all cells traversed by the linear approach path and loiter-down circle for collision \reviews{to obstacles}{review1:obstacles}{review1:approach_path} based on the elevation map and incorporating a safety margin.
%
%Disabling for now as > 8 pages..
\review{Note that approach path optimization, in this context, is only used for a more meaningful scoring of potential landing sites and the provision of an informed approach vector. It should not be seen as a replacement for local re-planners which are still necessary for real-time corrections upon the actual landing attempt.}{review1:landing}
%
%\timo{graphic of all decision layers}
%\subsection{Decision Making}
%
%The final decision making module considers slope and terrain roughness to return the safest landing spot within the ROI and a final validation grade
%
%For this, each layer is thresholded to obtain binary layers and the XOR of all layers is computed.
%
%We then apply distance transform to find the place furthest away from all hazards.
%
%\timo{MISSING: Incorporation of local wind field, nearby obstacles}
%
%\timo{graphic thresholded layers and XOR layer, distance transform and final landing spot}
%\subsection{Trajectory Generation and Landing Maneuver}
%
%Requires visual servoing towards designated landing spot. Compare descriptor-based and image/orthomosaic-based localization and visual servoing for this application.
%
%\clearpage
%
\subsection{Wind Vector Estimation}\label{sec:wind}
\review{The UAV's state estimator provides online estimates of the local wind field \cite{Leutenegger2014}.
All measurements taken within a certain distance to the center of a ROI are associated with a landing spot as shown in Fig. \ref{fig:rw_brazil}, denoted by the black circle.
To counteract slowly changing wind fields, we compute the final wind vector using the exponentially weighted moving average.}{review2:wind_vector}
\section{Results}
\subsection{Region Classification}
\label{ssec:exp-cla}
Fig. \ref{fig:roc_pr} shows the binary classification of homogenous regions into \verb|grass| or $\neg$\verb|grass|. 
The results from random forest \cite{Itseez2015}  and the multi-layer perceptron (MLP) based artificial neural network (ANN) \cite{Itseez2015} are plotted in form of the true positive rate (TPR) vs. the false positive rate (FPR) on the left (ROC chart) and the precision recall curve on the right.
While ANN performs slightly better in the validation set, the measured computational cost to predict a binary label is $2.4e-3 \pm \unit[7.8{e-4}]{ms}$ for RF and $1.7e-2 \pm \unit[9.0{e-3}]{ms}$ for ANN.
Since ROIs are tracked over several frames (cf. Sec.~\ref{sec:region_manager}) the influence of a single false prediction is mitigated by the probabilistic score as seen in Fig.~\ref{fig:synth} and Fig.~\ref{fig:rw_tobelhof}.
Hence RF was employed in all further experiments for computational speed-up.
%
%\timo{Show that the last sentence is correct}
\setlength\figureheight{1.4cm}
\setlength\figurewidth{2cm}
\begin{figure}[htb]
\centering
%\includegraphics[scale=1]{tikz/classification_legend.pdf} \\
%\hfill
\begin{subfigure}{0.4\linewidth}
\begin{tikzpicture}

\begin{axis}[%
width=0.951\figurewidth,
height=\figureheight,
at={(0\figurewidth,0\figureheight)},
scale only axis,
xmin=0,
xmax=1,
xtick={0,0.1,0.2,0.3,0.4,0.5,0.6,0.7,0.8,0.9,1},
xticklabels={{0},{},{},{},{},{0.5},{},{},{},{},{1}},
xlabel style={font=\color{white!15!black}},
xlabel={FPR},
ymin=0,
ymax=1,
ytick={0,0.1,0.2,0.3,0.4,0.5,0.6,0.7,0.8,0.9,1},
yticklabels={{0},{},{},{},{},{0.5},{},{},{},{},{1}},
ylabel style={font=\color{white!15!black}},
ylabel={TPR},
axis background/.style={fill=white},
axis x line*=bottom,
axis y line*=left,
xmajorgrids,
ymajorgrids
]
\addplot [color=blue, line width=1.0pt, forget plot]
  table[row sep=crcr]{%
0	0\\
0.0416666666666667	0.585365853658537\\
0.0416666666666667	0.585365853658537\\
0.0416666666666667	0.646341463414634\\
0.0520833333333333	0.707317073170732\\
0.0625	0.768292682926829\\
0.114583333333333	0.792682926829268\\
0.145833333333333	0.804878048780488\\
0.1875	0.853658536585366\\
0.21875	0.865853658536585\\
0.25	0.902439024390244\\
0.322916666666667	0.939024390243902\\
0.333333333333333	0.939024390243902\\
0.395833333333333	0.939024390243902\\
1	1\\
};
\addplot [color=red, line width=1.0pt, forget plot]
  table[row sep=crcr]{%
0	0\\
0	0.024390243902439\\
0	0.0487804878048781\\
0	0.0609756097560976\\
0	0.0853658536585366\\
0	0.109756097560976\\
0	0.134146341463415\\
0	0.146341463414634\\
0	0.182926829268293\\
0	0.195121951219512\\
0	0.219512195121951\\
0	0.24390243902439\\
0	0.25609756097561\\
0	0.280487804878049\\
0	0.304878048780488\\
0	0.329268292682927\\
0	0.341463414634146\\
0	0.365853658536585\\
0	0.390243902439024\\
0	0.414634146341463\\
0	0.439024390243902\\
0	0.451219512195122\\
0	0.475609756097561\\
0.0104166666666667	0.48780487804878\\
0.0104166666666667	0.51219512195122\\
0.0104166666666667	0.536585365853659\\
0.0104166666666667	0.548780487804878\\
0.0104166666666667	0.573170731707317\\
0.0104166666666667	0.597560975609756\\
0.0104166666666667	0.621951219512195\\
0.0208333333333333	0.621951219512195\\
0.0208333333333333	0.646341463414634\\
0.03125	0.658536585365854\\
0.03125	0.682926829268293\\
0.0416666666666667	0.695121951219512\\
0.0416666666666667	0.707317073170732\\
0.0520833333333333	0.719512195121951\\
0.0625	0.731707317073171\\
0.0729166666666667	0.74390243902439\\
0.0729166666666667	0.75609756097561\\
0.0833333333333333	0.768292682926829\\
0.0833333333333333	0.792682926829268\\
0.09375	0.804878048780488\\
0.09375	0.829268292682927\\
0.104166666666667	0.829268292682927\\
0.114583333333333	0.841463414634146\\
0.125	0.853658536585366\\
0.125	0.878048780487805\\
0.135416666666667	0.878048780487805\\
0.145833333333333	0.890243902439024\\
0.166666666666667	0.890243902439024\\
0.1875	0.890243902439024\\
0.208333333333333	0.890243902439024\\
0.21875	0.890243902439024\\
0.239583333333333	0.890243902439024\\
0.260416666666667	0.890243902439024\\
0.260416666666667	0.914634146341463\\
0.270833333333333	0.914634146341463\\
0.28125	0.926829268292683\\
0.291666666666667	0.939024390243902\\
0.302083333333333	0.951219512195122\\
0.322916666666667	0.951219512195122\\
0.333333333333333	0.951219512195122\\
0.354166666666667	0.951219512195122\\
0.375	0.951219512195122\\
0.395833333333333	0.951219512195122\\
0.40625	0.951219512195122\\
0.416666666666667	0.963414634146341\\
0.4375	0.963414634146341\\
0.458333333333333	0.963414634146341\\
0.479166666666667	0.963414634146341\\
0.479166666666667	0.975609756097561\\
0.5	0.975609756097561\\
0.520833333333333	0.975609756097561\\
0.541666666666667	0.975609756097561\\
0.5625	0.975609756097561\\
0.572916666666667	0.975609756097561\\
0.59375	0.975609756097561\\
0.614583333333333	0.975609756097561\\
0.635416666666667	0.975609756097561\\
0.635416666666667	0.98780487804878\\
0.65625	0.98780487804878\\
0.677083333333333	0.98780487804878\\
0.697916666666667	0.98780487804878\\
0.708333333333333	1\\
0.71875	1\\
0.739583333333333	1\\
0.760416666666667	1\\
0.78125	1\\
0.791666666666667	1\\
0.8125	1\\
0.833333333333333	1\\
0.854166666666667	1\\
0.875	1\\
0.885416666666667	1\\
0.90625	1\\
0.927083333333333	1\\
0.947916666666667	1\\
0.958333333333333	1\\
0.979166666666667	1\\
1	1\\
};
\end{axis}
\end{tikzpicture}%
%%\caption{Image}
%%\label{fig:image1}
%\vspace{-20pt}
\end{subfigure}
\hfill
\begin{subfigure}{0.4\linewidth}
\begin{tikzpicture}

\begin{axis}[%
width=0.951\figurewidth,
height=\figureheight,
at={(0\figurewidth,0\figureheight)},
scale only axis,
xmin=0,
xmax=1,
xtick={0,0.1,0.2,0.3,0.4,0.5,0.6,0.7,0.8,0.9,1},
xticklabels={{0},{},{},{},{},{0.5},{},{},{},{},{1}},
xlabel style={font=\color{white!15!black}},
xlabel={Recall},
ymin=0,
ymax=1,
ytick={0,0.1,0.2,0.3,0.4,0.5,0.6,0.7,0.8,0.9,1},
yticklabels={{0},{},{},{},{},{0.5},{},{},{},{},{1}},
ylabel style={font=\color{white!15!black}},
ylabel={Precision},
axis background/.style={fill=white},
axis x line*=bottom,
axis y line*=left,
xmajorgrids,
ymajorgrids
]
\addplot [color=blue, line width=1.0pt, forget plot]
  table[row sep=crcr]{%
0	1\\
0.585365853658537	0.923076923076923\\
0.585365853658537	0.923076923076923\\
0.646341463414634	0.929824561403509\\
0.707317073170732	0.920634920634921\\
0.768292682926829	0.91304347826087\\
0.792682926829268	0.855263157894737\\
0.804878048780488	0.825\\
0.853658536585366	0.795454545454545\\
0.865853658536585	0.771739130434783\\
0.902439024390244	0.755102040816326\\
0.939024390243902	0.712962962962963\\
0.939024390243902	0.706422018348624\\
0.939024390243902	0.669565217391304\\
1	0.460674157303371\\
};
\addplot [color=red, line width=1.0pt, forget plot]
  table[row sep=crcr]{%
0	1\\
0.024390243902439	1\\
0.0487804878048781	1\\
0.0609756097560976	1\\
0.0853658536585366	1\\
0.109756097560976	1\\
0.134146341463415	1\\
0.146341463414634	1\\
0.182926829268293	1\\
0.195121951219512	1\\
0.219512195121951	1\\
0.24390243902439	1\\
0.25609756097561	1\\
0.280487804878049	1\\
0.304878048780488	1\\
0.329268292682927	1\\
0.341463414634146	1\\
0.365853658536585	1\\
0.390243902439024	1\\
0.414634146341463	1\\
0.439024390243902	1\\
0.451219512195122	1\\
0.475609756097561	1\\
0.48780487804878	0.975609756097561\\
0.51219512195122	0.976744186046512\\
0.536585365853659	0.977777777777778\\
0.548780487804878	0.978260869565217\\
0.573170731707317	0.979166666666667\\
0.597560975609756	0.98\\
0.621951219512195	0.980769230769231\\
0.621951219512195	0.962264150943396\\
0.646341463414634	0.963636363636364\\
0.658536585365854	0.947368421052632\\
0.682926829268293	0.949152542372881\\
0.695121951219512	0.934426229508197\\
0.707317073170732	0.935483870967742\\
0.719512195121951	0.921875\\
0.731707317073171	0.909090909090909\\
0.74390243902439	0.897058823529412\\
0.75609756097561	0.898550724637681\\
0.768292682926829	0.887323943661972\\
0.792682926829268	0.89041095890411\\
0.804878048780488	0.88\\
0.829268292682927	0.883116883116883\\
0.829268292682927	0.871794871794872\\
0.841463414634146	0.8625\\
0.853658536585366	0.853658536585366\\
0.878048780487805	0.857142857142857\\
0.878048780487805	0.847058823529412\\
0.890243902439024	0.839080459770115\\
0.890243902439024	0.820224719101124\\
0.890243902439024	0.802197802197802\\
0.890243902439024	0.78494623655914\\
0.890243902439024	0.776595744680851\\
0.890243902439024	0.760416666666667\\
0.890243902439024	0.744897959183674\\
0.914634146341463	0.75\\
0.914634146341463	0.742574257425743\\
0.926829268292683	0.737864077669903\\
0.939024390243902	0.733333333333333\\
0.951219512195122	0.728971962616822\\
0.951219512195122	0.715596330275229\\
0.951219512195122	0.709090909090909\\
0.951219512195122	0.696428571428571\\
0.951219512195122	0.684210526315789\\
0.951219512195122	0.672413793103448\\
0.951219512195122	0.666666666666667\\
0.963414634146341	0.663865546218487\\
0.963414634146341	0.652892561983471\\
0.963414634146341	0.642276422764228\\
0.963414634146341	0.632\\
0.975609756097561	0.634920634920635\\
0.975609756097561	0.625\\
0.975609756097561	0.615384615384615\\
0.975609756097561	0.606060606060606\\
0.975609756097561	0.597014925373134\\
0.975609756097561	0.592592592592593\\
0.975609756097561	0.583941605839416\\
0.975609756097561	0.575539568345324\\
0.975609756097561	0.567375886524823\\
0.98780487804878	0.570422535211268\\
0.98780487804878	0.5625\\
0.98780487804878	0.554794520547945\\
0.98780487804878	0.547297297297297\\
1	0.546666666666667\\
1	0.543046357615894\\
1	0.535947712418301\\
1	0.529032258064516\\
1	0.522292993630573\\
1	0.518987341772152\\
1	0.5125\\
1	0.506172839506173\\
1	0.5\\
1	0.493975903614458\\
1	0.491017964071856\\
1	0.485207100591716\\
1	0.47953216374269\\
1	0.473988439306358\\
1	0.471264367816092\\
1	0.465909090909091\\
1	0.460674157303371\\
};
\end{axis}
\end{tikzpicture}% 
%\caption{Image}
%\label{fig:image2}
\end{subfigure}
\hfill
\begin{subfigure}{0.13\linewidth}
\includegraphics[scale=0.7]{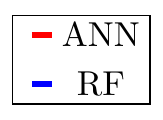}{} 
\end{subfigure}
\cprotect\caption{Binary classification of homogenous regions into \verb|grass| or $\neg$\verb|grass| for artificial neural network (ANN) and random forest (RF)}
%\timo{This needs more tuning.. (and/or more data)}}
\label{fig:roc_pr}
\end{figure}
%
%\subsection{Region merging}
%
\subsection{Computational Costs}
The runtime evaluated on the real-world experiment ``Switzerland" is shown in Table~\ref{tab:runtime}.
In the frontend, most of the time is spent on classification, in particular, to compute Gabor features.
The frontend can run at $\unit[12.49]{Hz}$ with Gabor features and at $\unit[21.82]{Hz}$ when only relying on color cues.
One could speed up the Gabor filter by only retrieving few samples from the image patch instead of using a convolution over the whole patch.
However, since the incoming image rate is $\unit[4]{Hz}$, the frontend (including Gabor features) is more than three times faster than real-time.
Note the efficiency of the geometric region managing.
\review{BA, dense reconstruction and terrain analysis introduce, depending on the grid resolution, a certain delay and are available in \emph{near real-time}.}{review1:real_time}
\begin{table}[!ht]
            \scriptsize
    \begin{tabular}[b]{|l|l|l|}\hline
 & samples & mean $\pm$ stdev \\ \hline
\rowcolor{black!10} Segmentation & $250$ & $7.20\pm 0.64$  \\\hline
\rowcolor{black!10} Class. w. Gabor & $250$ &$70.30\pm22.76$  \\
\rowcolor{black!10} - feature vector & $2217$ &$3.56\pm 4.95$  \\
\rowcolor{black!10} - predict & $2217$ &$2.4e{-3}\pm 7.8e{-4}$  \\\hline
\rowcolor{black!10} Class. w/o Gabor & $250$ &$36.89\pm20.28$  \\
\rowcolor{black!10} - feature vector & $2217$ &$0.21\pm0.27$  \\
\rowcolor{black!10} - predict & $2217$ &$2.1e{-3}\pm8.4e{-4}$  \\\hline
\rowcolor{black!10} Region Manager &$250$ &$1.57\pm0.39$  \\\hline
\rowcolor{black!30} Bundle Adjustm. & $23$ & $20.1\pm 3.2$  \\\hline
\rowcolor{black!30} Dense Reconstr. & $23$  &$16.7\pm 2.3$  \\\hline
\rowcolor{black!30} Decision Layers & $10$ &$1083\pm 21.09$  \\\hline
\rowcolor{black!30} Approach Vector & $10$ &$527.48 \pm 20.65$  \\\hline 
\end{tabular}
\includegraphics[width=2.45cm]{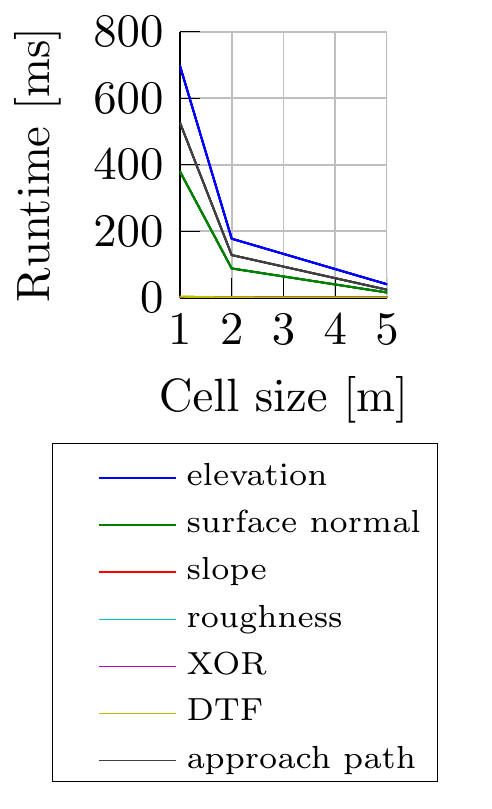}
\vspace{10pt}
\caption[]{Runtime in $ms$ \review{for the ``Switzerland" dataset}{review2:runtime}. Note that the frontend (light gray) and backend (dark gray) run in separate threads. Runtime of frontend, BA, and dense reconstruction is measured per frame. Runtime of decision layers and approach vectors is computed for a ROI point cloud consisting of $\num{4.2e6}$ points ($300\times 300$ cells, $\unit[1.0]{m}$ resolution). Evaluated on Intel(R) Core(TM) i7-4800MQ CPU @ 2.70GHz.}
%    \captionlistentry[table]{A table beside a figure}
%    \captionsetup{labelformat=andtable}
%\caption
\vspace{-10pt}
\label{tab:runtime}
  \end{table}
\subsection{Semi-Synthetic Dataset}
\begin{figure*}[htb]
 % ################# 1st row
  %  \subcaptionbox{Overview mesh}{% 
      \includegraphics[width=\linewidth]{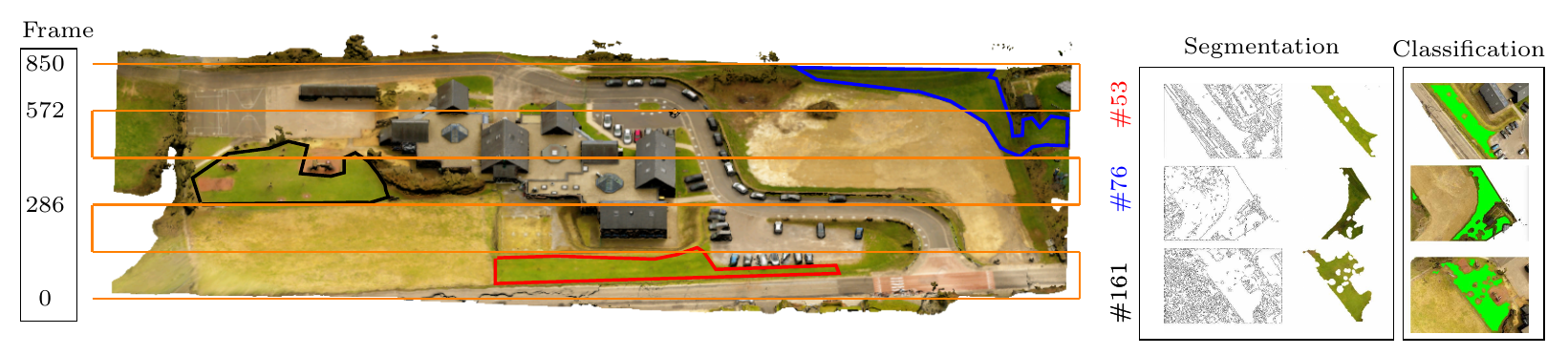}% 
  %  }
    \\
    % ################# 2nd row
    \raisebox{-10pt}{\scalebox{0.75}{\input{matlab/synthetic/roi_analysis_observations_v2.tikz}}}
    \raisebox{-10pt}{\scalebox{0.75}{\input{matlab/synthetic/roi_analysis_rate_v2.tikz}}}
   \raisebox{-10pt}{\scalebox{0.75}{\input{matlab/synthetic/roi_analysis_area_v2.tikz}}}
    \raisebox{-10pt}{\scalebox{0.75}{\input{matlab/synthetic/roi_analysis_score_v2.tikz}}}
    \subcaptionbox{Bundle adjustment\label{fig:synth_a}}{% 
      \includegraphics[width=0.18\linewidth]{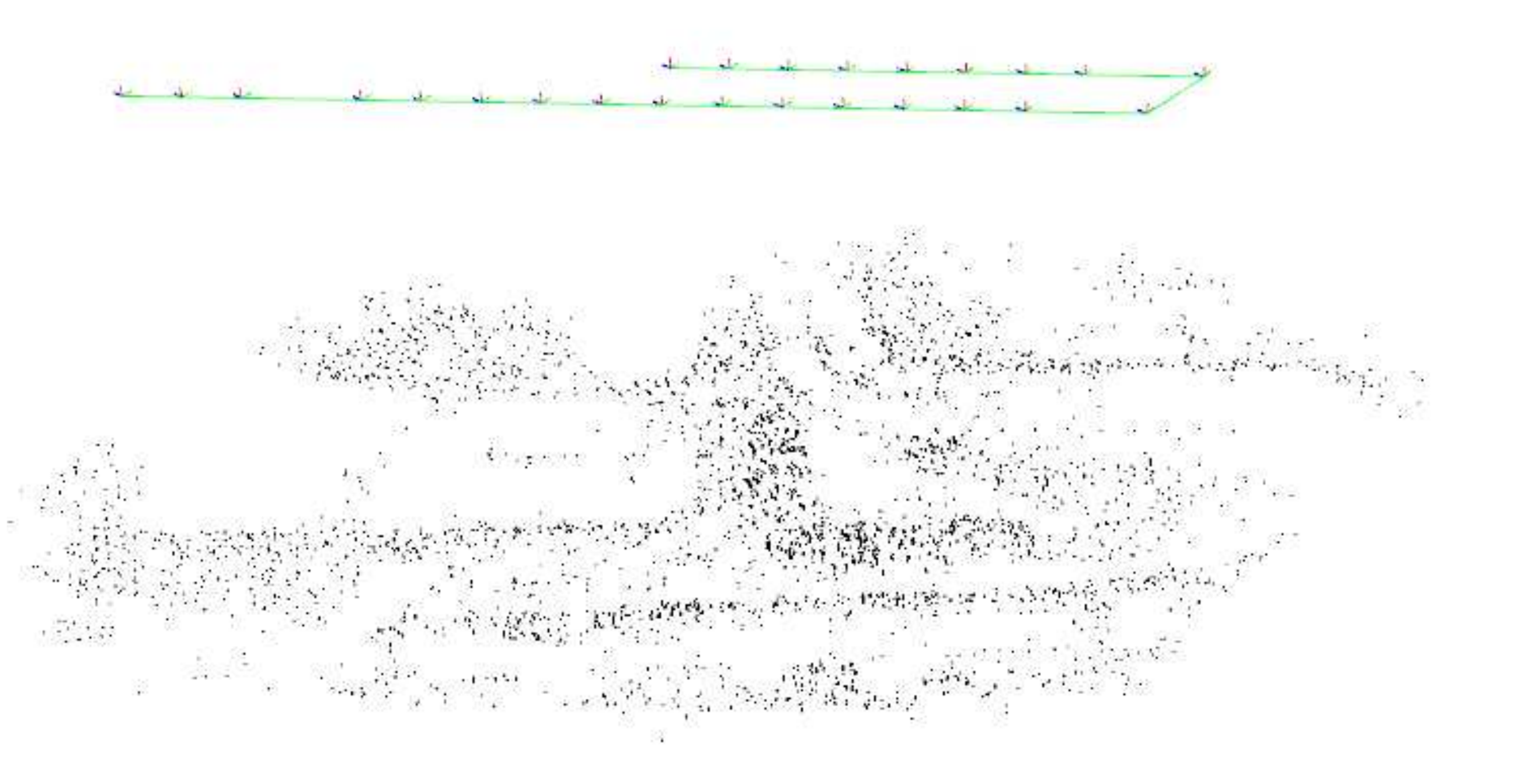}% 
    } 
            \hfill
    \subcaptionbox{Dense reconstruction\label{fig:synth_b}}{% 
      \includegraphics[width=0.18\linewidth]{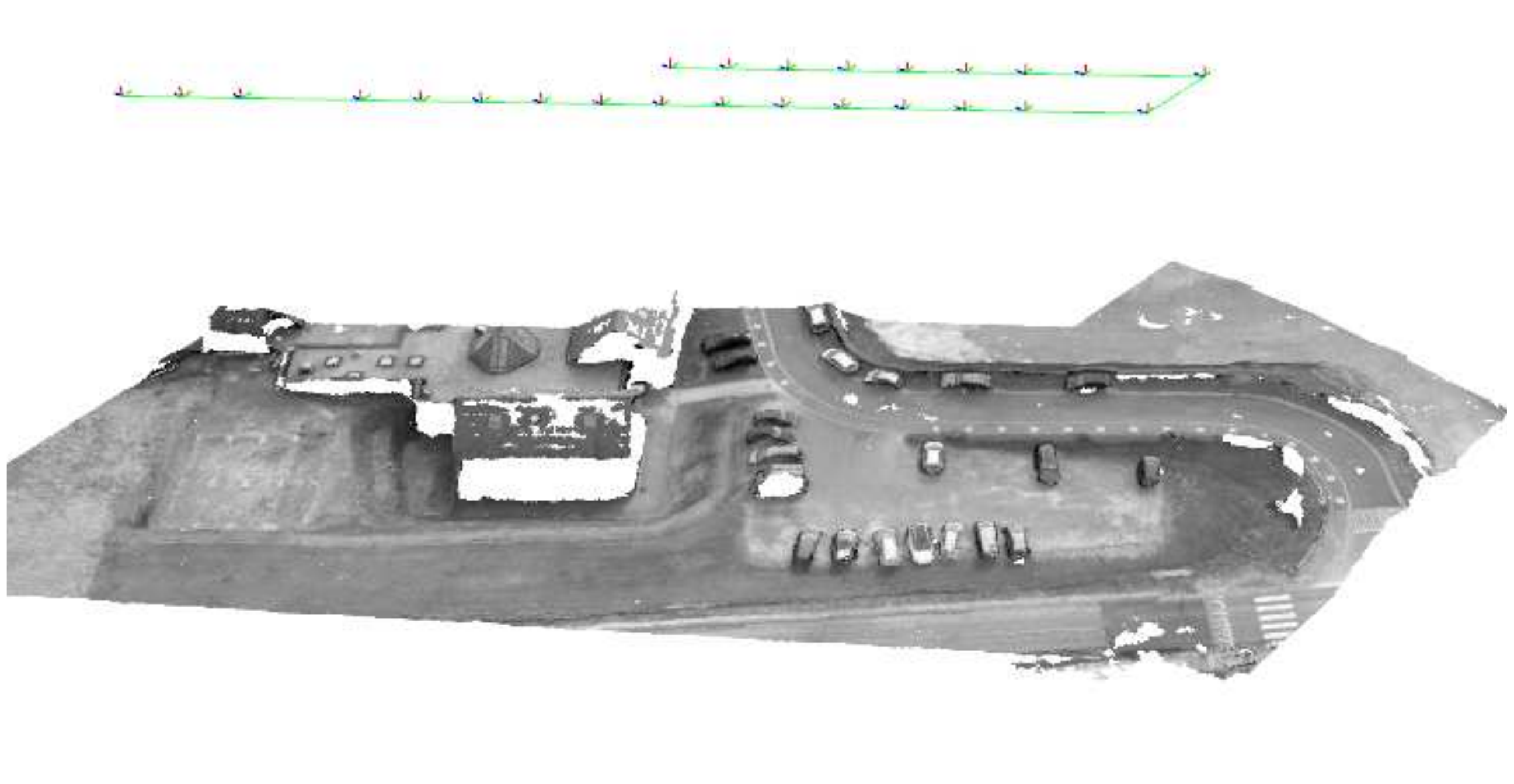}% 
    } 
    \hfill
    % ################# 3rd row
    \\
    \begin{minipage}[t]{0.14\textwidth}
    \subcaptionbox{$2.5$D elevation\label{fig:synth_c}}{% 
     \includegraphics[width=1\linewidth]{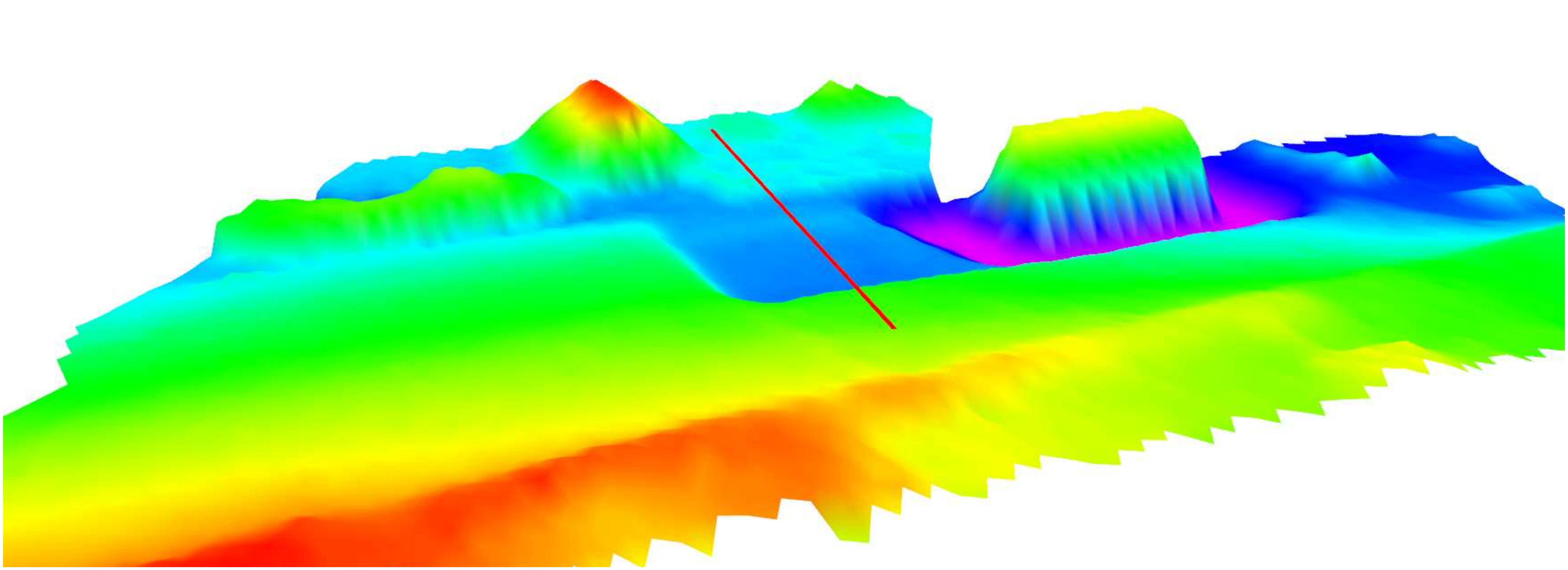}% 
     }
     \end{minipage}
    	\hfill
    	\begin{minipage}[t]{0.14\textwidth}
        \subcaptionbox{Slope\label{fig:synth_d}}{% 
\includegraphics[width=1\linewidth]{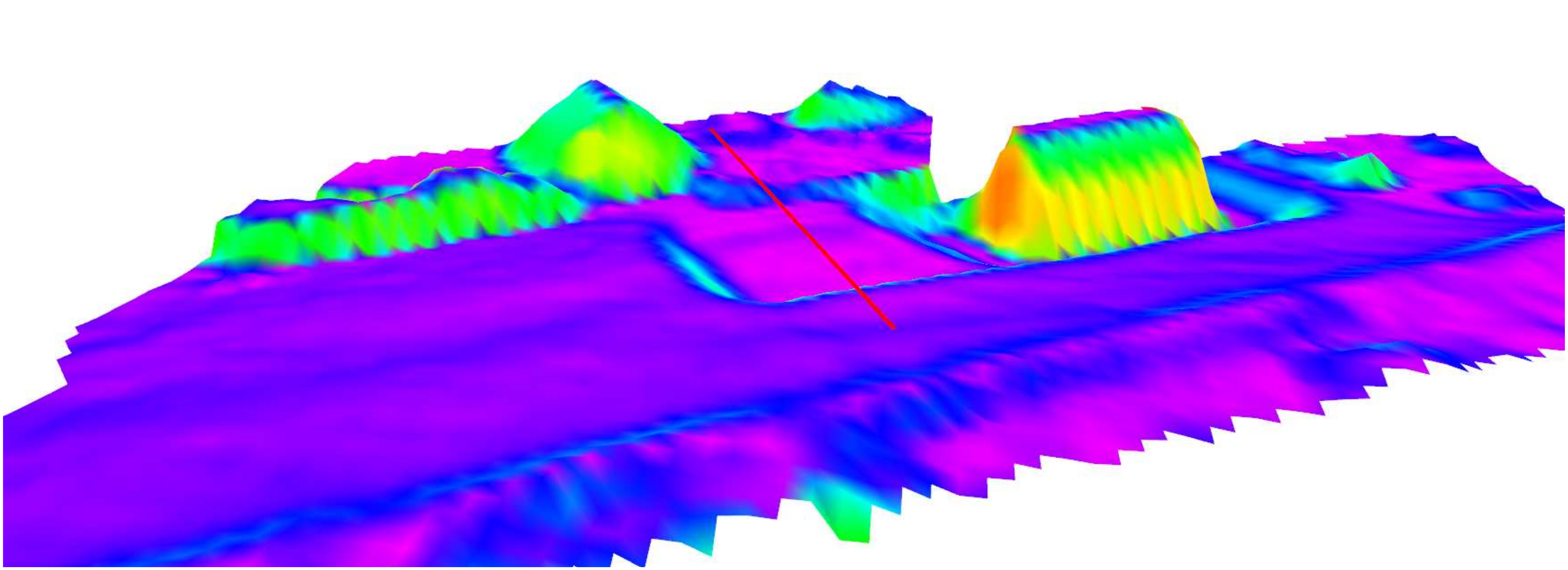}% 
    } 
    \end{minipage}
    	\hfill
    	\begin{minipage}[t]{0.14\textwidth}
           \subcaptionbox{Roughness\label{fig:synth_e}}{% 
\includegraphics[width=1\linewidth]{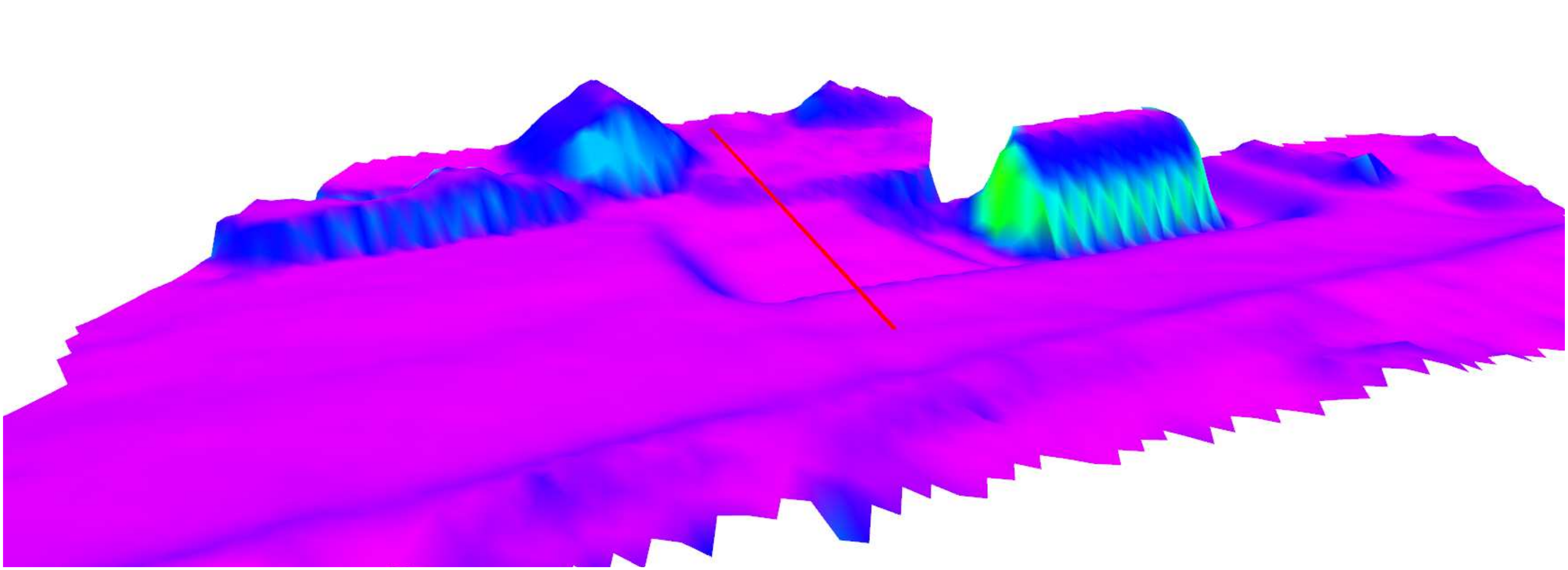}% 
    }
    \end{minipage}
    	\hfill
    	\begin{minipage}[t]{0.14\textwidth}
            \subcaptionbox{Bin. roughness \label{fig:synth_f}}{% 
\includegraphics[width=1\linewidth]{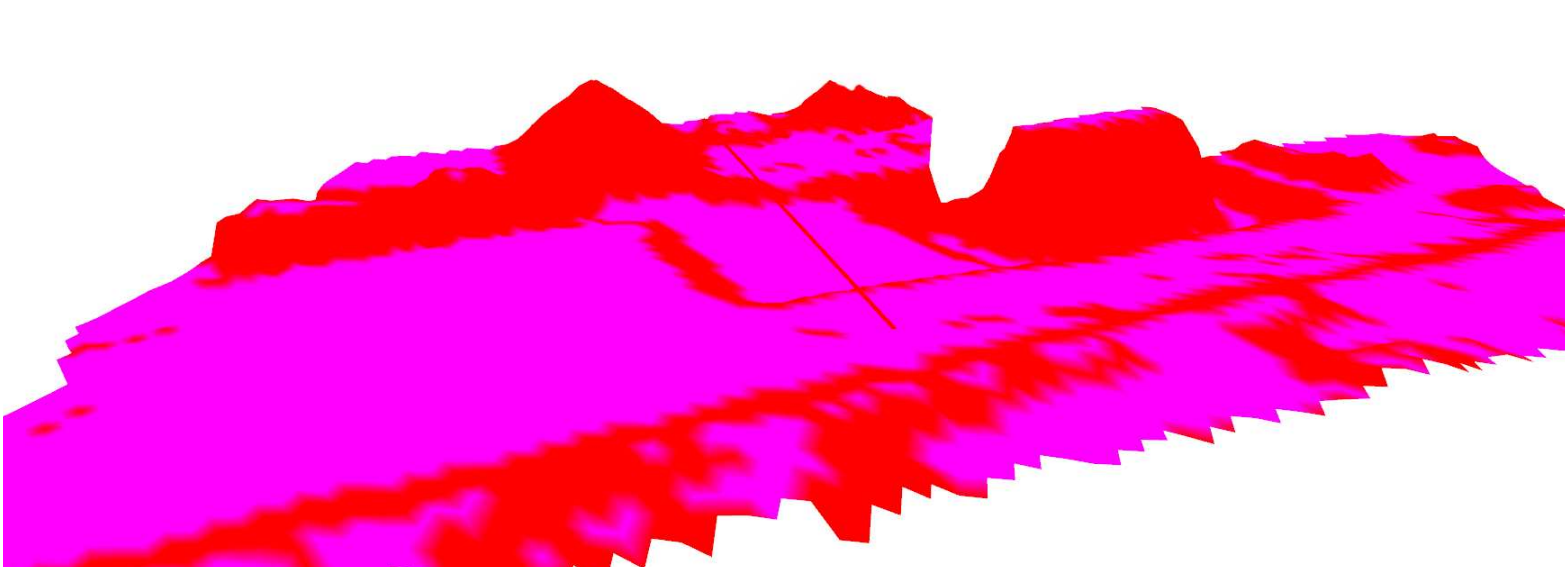}% 
     }
     \end{minipage}
    	\hfill
    	\begin{minipage}[t]{0.33\linewidth}
           \subcaptionbox{Distance map (side, top) for ROI $\#53$ \label{fig:synth_g}}{% 
\includegraphics[width=0.9\linewidth, trim = 0 0 0 50,clip=true]{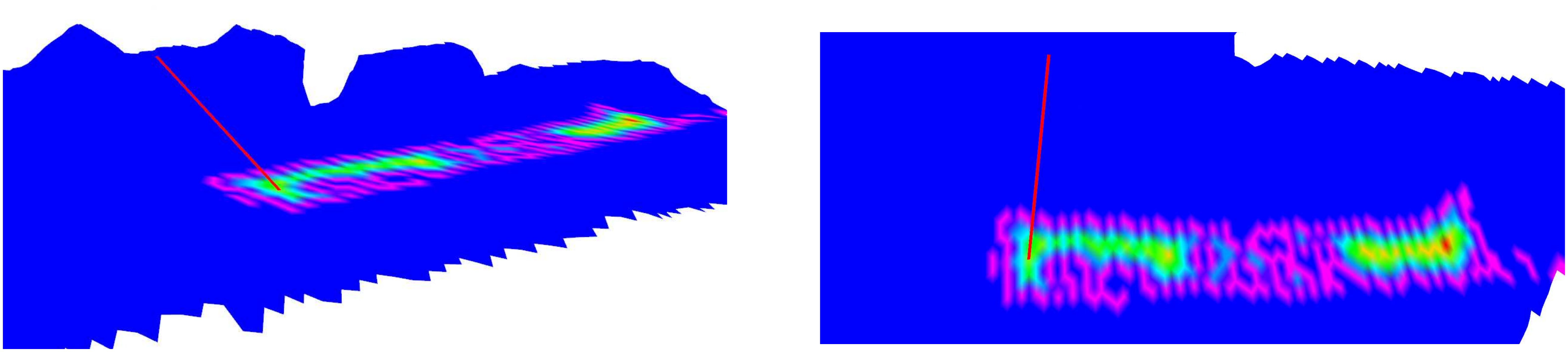}% 
    }  \end{minipage}
    \caption[]{Semi-synthetic dataset illustrating the output of the segmentation, classification, tracking over time, and fine $3$D terrain evaluation. The simulated camera is a down-looking \textit{Aptina MT9v034} ($\unit[0.36]{MP}$). The (unscaled) mesh was downloaded from \url{https://skfb.ly/6o9Y7}.}
    \label{fig:synth}
  \end{figure*}

%
%\begin{figure}[b!]
%\centering\includegraphics[width=1\textwidth]{tikz/results_decision_synthetic_v2}
%\caption{Simulation, Backend: dense reconstruction, and computation of a valid approach path.}
%\label{fig:synth_backend}
%\end{figure}
%
The results obtained from a synthetic dataset are shown in Fig.~\ref{fig:synth}.
The images are rendered using \textit{Blender} from poses computed by a simple lawn mower scan pattern generator.
The underlying mesh was obtained from photogrammetry with images taken from a real camera, hence denoted as \textit{semi}-synthetic.
The overview mesh in Fig.~\ref{fig:synth} shows the scan pattern, corresponding frame indices on the left, and the three landing spots with the highest score.
The right side of Fig.~\ref{fig:synth} shows the output of segmentation and classification for a sample frame.
%
%#########################################################
The second row of Fig.~\ref{fig:synth} plots the number of observations, classification certainty, score, and estimated area over time.
Although there are some wrong classifications, the final classification \review{certainty}{review2:certainty} for all three regions is above $\unit[95]{\%}$.
Fig.~\ref{fig:synth_a} to \ref{fig:synth_g} show the backend, i.e. the dense reconstruction, decision, and hazard layers, the final touch down point, and linear approach path for the highest-scoring ROI $\#53$.
Note that the touch down points on the right side of Fig \ref{fig:synth_g} show a high distance to the next hazards within the ROI but are rejected due to obstacles (house) along the planned linear approach path.
\subsection{Experiments with Real-World Datasets}\label{rw_experiments}
\begin{figure*}[htb]
\begin{center}
\begin{minipage}[t]{0.19\textwidth}
\centering
    \subcaptionbox{Overview mesh\label{fig:rw_tobelhof_a}}{% 
    \centering
      \includegraphics[width=1\linewidth]{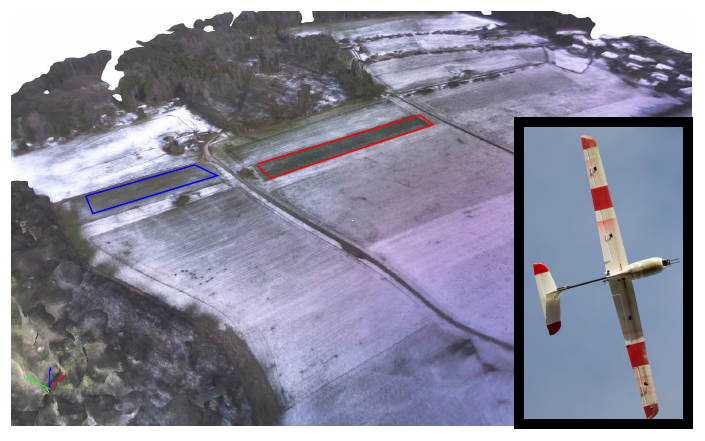}% 
    }
    \end{minipage}
    \hfill
    \begin{minipage}[t]{0.19\textwidth}
    \centering
    \subcaptionbox{Segmentation\label{fig:rw_tobelhof_b}}{% 
    \centering
      \includegraphics[width=1\linewidth]{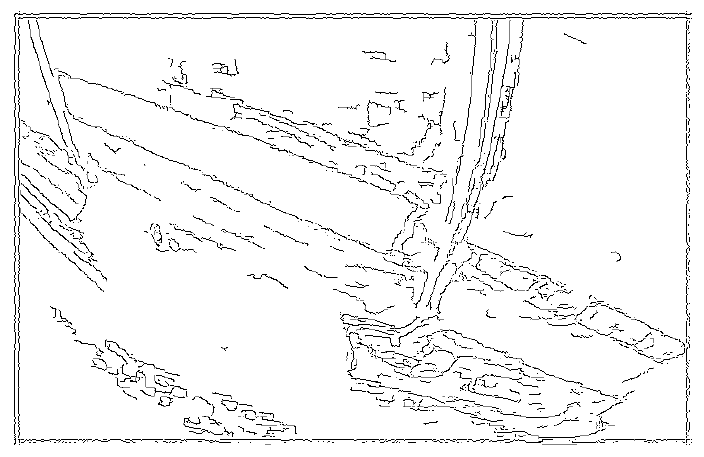}% 
    }
    \end{minipage}
        \hfill
	\begin{minipage}[t]{0.19\textwidth}
    %\subcaptionbox{Classification certainty}{% 
    \centering
    \raisebox{-10pt}{\scalebox{0.8}{\input{matlab/tobelhof/roi_analysis_rate_v2.tikz}}}% 
    \end{minipage}
    %} 
    %
        \hfill
    \begin{minipage}[t]{0.2\textwidth}
    \centering
    \subcaptionbox{$2.5$D elevation map}{% 
        \centering
       \includegraphics[width=1\linewidth]{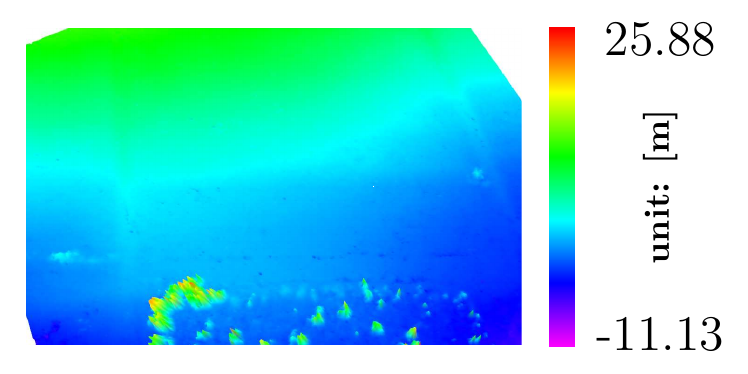}% 
    } 
    \end{minipage}
        \hfill
    \begin{minipage}[t]{0.2\textwidth}
    \centering
    \subcaptionbox{Terrain slope }{% 
      \centering\includegraphics[width=1\linewidth]{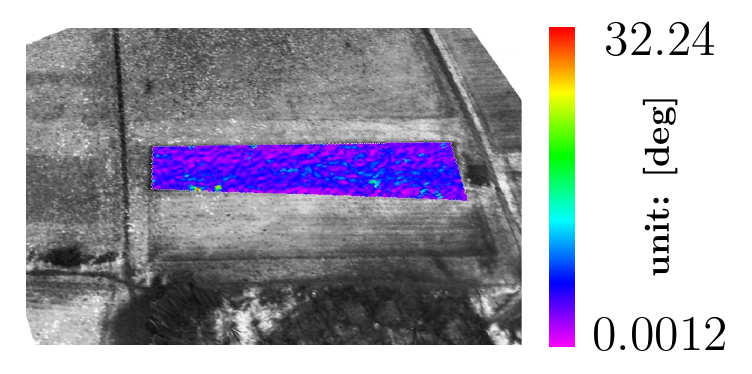}% 
    } 
    \end{minipage}
    \\
    \begin{minipage}[t]{0.19\textwidth}
    \centering
    \subcaptionbox{Classification \label{fig:rw_tobelhof_e}}{% 
    \centering
      \includegraphics[width=1\linewidth]{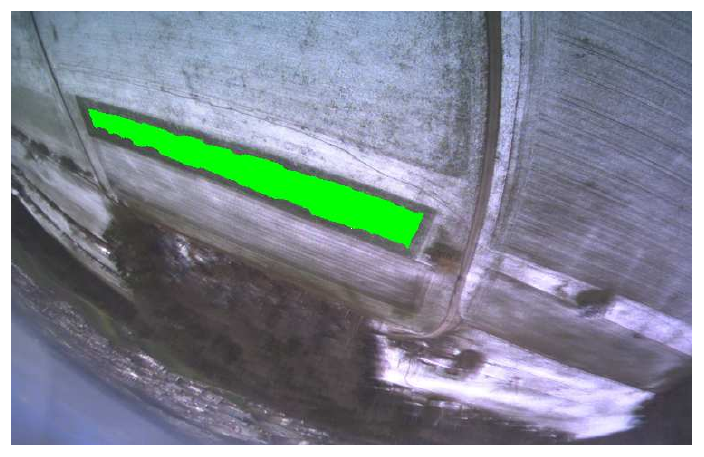}% 
    } 
    \end{minipage}
        \hfill
        \begin{minipage}[t]{0.19\textwidth}
        \centering
    %    \subcaptionbox{Number observations}{% 
      \raisebox{-10pt}{\scalebox{0.8}{% This file was created by matlab2tikz.
\begin{tikzpicture}

\begin{axis}[%
width=0.951\figurewidth,
height=\figureheight,
at={(0\figurewidth,0\figureheight)},
scale only axis,
separate axis lines,
every outer x axis line/.append style={black},
every x tick label/.append style={font=\color{black}},
every x tick/.append style={black},
xmin=0,
xmax=300,
xlabel={Frame},
every outer y axis line/.append style={black},
every y tick label/.append style={font=\color{black}},
every y tick/.append style={black},
ymin=0,
ymax=80,
ylabel={Num. obs.},
axis background/.style={fill=white},
xmajorgrids,
xminorgrids,
ymajorgrids,
yminorgrids
]
\addplot [color=red, line width=1.2pt, forget plot]
  table[row sep=crcr]{%
1	1\\
2	2\\
3	3\\
4	4\\
5	5\\
6	6\\
7	7\\
8	8\\
9	9\\
10	10\\
11	11\\
12	12\\
13	13\\
14	13\\
15	13\\
16	13\\
17	13\\
18	13\\
19	13\\
20	13\\
21	13\\
22	13\\
23	13\\
24	13\\
25	13\\
26	13\\
27	13\\
28	13\\
29	13\\
30	13\\
31	13\\
32	13\\
33	13\\
34	13\\
35	13\\
36	13\\
37	13\\
38	13\\
39	13\\
40	13\\
41	13\\
42	13\\
43	13\\
44	13\\
45	13\\
46	13\\
47	13\\
48	13\\
49	13\\
50	13\\
51	13\\
52	13\\
53	13\\
54	13\\
55	13\\
56	13\\
57	13\\
58	13\\
59	13\\
60	13\\
61	13\\
62	13\\
63	13\\
64	13\\
65	13\\
66	13\\
67	13\\
68	13\\
69	13\\
70	13\\
71	13\\
72	13\\
73	13\\
74	13\\
75	13\\
76	13\\
77	13\\
78	13\\
79	13\\
80	13\\
81	13\\
82	13\\
83	13\\
84	13\\
85	13\\
86	13\\
87	13\\
88	13\\
89	13\\
90	13\\
91	13\\
92	13\\
93	13\\
94	13\\
95	13\\
96	13\\
97	13\\
98	13\\
99	13\\
100	13\\
101	13\\
102	13\\
103	13\\
104	13\\
105	13\\
106	13\\
107	13\\
108	13\\
109	13\\
110	13\\
111	13\\
112	13\\
113	13\\
114	13\\
115	13\\
116	13\\
117	13\\
118	13\\
119	13\\
120	13\\
121	13\\
122	13\\
123	13\\
124	13\\
125	13\\
126	13\\
127	13\\
128	13\\
129	13\\
130	13\\
131	13\\
132	13\\
133	13\\
134	13\\
135	13\\
136	13\\
137	13\\
138	13\\
139	13\\
140	13\\
141	13\\
142	13\\
143	13\\
144	13\\
145	13\\
146	13\\
147	13\\
148	13\\
149	14\\
150	15\\
151	15\\
152	16\\
153	18\\
154	19\\
155	19\\
156	19\\
157	21\\
158	22\\
159	22\\
160	24\\
161	25\\
162	26\\
163	27\\
164	28\\
165	29\\
166	30\\
167	30\\
168	31\\
169	32\\
170	34\\
171	35\\
172	35\\
173	36\\
174	37\\
175	38\\
176	39\\
177	40\\
178	41\\
179	42\\
180	43\\
181	44\\
182	45\\
183	46\\
184	47\\
185	47\\
186	47\\
187	47\\
188	48\\
189	49\\
190	49\\
191	49\\
192	49\\
193	49\\
194	49\\
195	49\\
196	49\\
197	49\\
198	49\\
199	49\\
200	49\\
201	49\\
202	49\\
203	49\\
204	49\\
205	49\\
206	49\\
207	49\\
208	49\\
209	49\\
210	49\\
211	49\\
212	49\\
213	49\\
214	49\\
215	49\\
216	49\\
217	49\\
218	49\\
219	49\\
220	49\\
221	49\\
222	49\\
223	49\\
224	49\\
225	49\\
226	49\\
227	49\\
228	49\\
229	49\\
230	49\\
231	49\\
232	49\\
233	49\\
234	49\\
235	49\\
236	49\\
237	49\\
238	49\\
239	49\\
240	49\\
241	49\\
242	49\\
243	49\\
244	49\\
245	49\\
246	49\\
247	49\\
248	49\\
249	49\\
250	49\\
251	49\\
252	49\\
253	49\\
254	49\\
255	49\\
256	49\\
257	49\\
258	49\\
259	49\\
260	49\\
261	49\\
262	49\\
263	49\\
264	49\\
265	49\\
266	49\\
267	49\\
268	49\\
269	49\\
270	49\\
271	49\\
272	49\\
273	49\\
274	49\\
275	49\\
276	49\\
277	49\\
278	49\\
279	49\\
280	49\\
281	49\\
282	49\\
283	49\\
284	49\\
285	49\\
286	49\\
287	49\\
288	49\\
289	49\\
};
\addplot [color=blue, line width=1.2pt, forget plot]
  table[row sep=crcr]{%
1	0\\
2	0\\
3	0\\
4	0\\
5	0\\
6	0\\
7	0\\
8	0\\
9	0\\
10	0\\
11	0\\
12	0\\
13	0\\
14	0\\
15	0\\
16	0\\
17	0\\
18	0\\
19	0\\
20	0\\
21	0\\
22	0\\
23	0\\
24	0\\
25	0\\
26	1\\
27	2\\
28	3\\
29	4\\
30	5\\
31	5\\
32	5\\
33	5\\
34	5\\
35	5\\
36	5\\
37	5\\
38	5\\
39	5\\
40	5\\
41	5\\
42	5\\
43	5\\
44	5\\
45	5\\
46	5\\
47	5\\
48	5\\
49	5\\
50	5\\
51	5\\
52	5\\
53	5\\
54	5\\
55	5\\
56	5\\
57	5\\
58	5\\
59	5\\
60	5\\
61	5\\
62	5\\
63	5\\
64	5\\
65	5\\
66	5\\
67	5\\
68	5\\
69	5\\
70	5\\
71	5\\
72	5\\
73	5\\
74	5\\
75	5\\
76	5\\
77	5\\
78	5\\
79	5\\
80	5\\
81	5\\
82	5\\
83	5\\
84	5\\
85	5\\
86	5\\
87	5\\
88	5\\
89	5\\
90	5\\
91	5\\
92	5\\
93	5\\
94	5\\
95	5\\
96	5\\
97	5\\
98	5\\
99	5\\
100	5\\
101	5\\
102	5\\
103	5\\
104	5\\
105	5\\
106	5\\
107	5\\
108	5\\
109	5\\
110	5\\
111	5\\
112	5\\
113	5\\
114	5\\
115	5\\
116	5\\
117	5\\
118	5\\
119	5\\
120	5\\
121	5\\
122	5\\
123	5\\
124	5\\
125	5\\
126	5\\
127	6\\
128	6\\
129	6\\
130	6\\
131	8\\
132	8\\
133	8\\
134	9\\
135	9\\
136	9\\
137	10\\
138	10\\
139	11\\
140	12\\
141	13\\
142	14\\
143	15\\
144	16\\
145	17\\
146	18\\
147	19\\
148	20\\
149	21\\
150	22\\
151	23\\
152	24\\
153	25\\
154	26\\
155	27\\
156	28\\
157	29\\
158	30\\
159	31\\
160	31\\
161	32\\
162	32\\
163	33\\
164	34\\
165	35\\
166	36\\
167	37\\
168	38\\
169	39\\
170	40\\
171	40\\
172	41\\
173	41\\
174	42\\
175	43\\
176	44\\
177	45\\
178	46\\
179	47\\
180	48\\
181	49\\
182	50\\
183	51\\
184	52\\
185	53\\
186	54\\
187	55\\
188	56\\
189	57\\
190	58\\
191	59\\
192	60\\
193	61\\
194	62\\
195	63\\
196	64\\
197	65\\
198	66\\
199	67\\
200	68\\
201	69\\
202	70\\
203	71\\
204	72\\
205	73\\
206	74\\
207	75\\
208	76\\
209	77\\
210	77\\
211	77\\
212	77\\
213	77\\
214	77\\
215	77\\
216	77\\
217	77\\
218	77\\
219	77\\
220	77\\
221	77\\
222	77\\
223	77\\
224	77\\
225	77\\
226	77\\
227	77\\
228	77\\
229	77\\
230	77\\
231	77\\
232	77\\
233	77\\
234	77\\
235	77\\
236	77\\
237	77\\
238	77\\
239	77\\
240	77\\
241	77\\
242	77\\
243	77\\
244	77\\
245	77\\
246	77\\
247	77\\
248	77\\
249	77\\
250	77\\
251	77\\
252	77\\
253	77\\
254	77\\
255	77\\
256	77\\
257	77\\
258	77\\
259	77\\
260	77\\
261	77\\
262	77\\
263	77\\
264	77\\
265	77\\
266	77\\
267	77\\
268	77\\
269	77\\
270	77\\
271	77\\
272	77\\
273	77\\
274	77\\
275	77\\
276	77\\
277	77\\
278	77\\
279	77\\
280	77\\
281	77\\
282	77\\
283	77\\
284	77\\
285	77\\
286	77\\
287	77\\
288	77\\
289	77\\
};
\end{axis}
\end{tikzpicture}%
}}
    %} 
    \end{minipage}
        \hfill
        \begin{minipage}[t]{0.19\textwidth}
        \centering
       % \subcaptionbox{ROI score}{% 
     \raisebox{-10pt}{\scalebox{0.8}{% This file was created by matlab2tikz.
\begin{tikzpicture}

\begin{axis}[%
width=0.951\figurewidth,
height=\figureheight,
at={(0\figurewidth,0\figureheight)},
scale only axis,
separate axis lines,
every outer x axis line/.append style={black},
every x tick label/.append style={font=\color{black}},
every x tick/.append style={black},
xmin=0,
xmax=300,
xlabel={Frame},
every outer y axis line/.append style={black},
every y tick label/.append style={font=\color{black}},
every y tick/.append style={black},
ymin=-20,
ymax=100,
ylabel={Score [\%]},
axis background/.style={fill=white},
xmajorgrids,
xminorgrids,
ymajorgrids,
yminorgrids
]
\addplot [color=red, line width=1.2pt, forget plot]
  table[row sep=crcr]{%
1	-19\\
2	-18\\
3	-17\\
4	-16\\
5	-15\\
6	-14\\
7	-13\\
8	-12\\
9	-11\\
10	80\\
11	81\\
12	83\\
13	76\\
14	76\\
15	76\\
16	76\\
17	76\\
18	76\\
19	76\\
20	76\\
21	76\\
22	76\\
23	76\\
24	76\\
25	76\\
26	76\\
27	76\\
28	76\\
29	76\\
30	76\\
31	76\\
32	76\\
33	76\\
34	76\\
35	76\\
36	76\\
37	76\\
38	76\\
39	76\\
40	76\\
41	76\\
42	76\\
43	76\\
44	76\\
45	76\\
46	76\\
47	76\\
48	76\\
49	76\\
50	76\\
51	76\\
52	76\\
53	76\\
54	76\\
55	76\\
56	76\\
57	76\\
58	76\\
59	76\\
60	76\\
61	76\\
62	76\\
63	76\\
64	76\\
65	76\\
66	76\\
67	76\\
68	76\\
69	76\\
70	76\\
71	76\\
72	76\\
73	76\\
74	76\\
75	76\\
76	76\\
77	76\\
78	76\\
79	76\\
80	76\\
81	76\\
82	76\\
83	76\\
84	76\\
85	76\\
86	76\\
87	76\\
88	76\\
89	76\\
90	76\\
91	76\\
92	76\\
93	76\\
94	76\\
95	76\\
96	76\\
97	76\\
98	76\\
99	76\\
100	76\\
101	76\\
102	76\\
103	76\\
104	76\\
105	76\\
106	76\\
107	76\\
108	76\\
109	76\\
110	76\\
111	76\\
112	76\\
113	76\\
114	76\\
115	76\\
116	76\\
117	76\\
118	76\\
119	76\\
120	76\\
121	76\\
122	76\\
123	76\\
124	76\\
125	76\\
126	76\\
127	76\\
128	76\\
129	76\\
130	76\\
131	76\\
132	76\\
133	76\\
134	76\\
135	76\\
136	76\\
137	76\\
138	76\\
139	76\\
140	76\\
141	76\\
142	76\\
143	76\\
144	76\\
145	76\\
146	76\\
147	76\\
148	76\\
149	78\\
150	80\\
151	80\\
152	75\\
153	77\\
154	78\\
155	78\\
156	78\\
157	80\\
158	77\\
159	77\\
160	79\\
161	80\\
162	80\\
163	81\\
164	82\\
165	82\\
166	83\\
167	83\\
168	83\\
169	84\\
170	82\\
171	80\\
172	80\\
173	80\\
174	78\\
175	78\\
176	79\\
177	80\\
178	80\\
179	80\\
180	81\\
181	81\\
182	82\\
183	82\\
184	82\\
185	82\\
186	82\\
187	82\\
188	81\\
189	79\\
190	79\\
191	79\\
192	79\\
193	79\\
194	79\\
195	79\\
196	79\\
197	79\\
198	79\\
199	79\\
200	79\\
201	79\\
202	79\\
203	79\\
204	79\\
205	79\\
206	79\\
207	79\\
208	79\\
209	79\\
210	79\\
211	79\\
212	79\\
213	79\\
214	79\\
215	79\\
216	79\\
217	79\\
218	79\\
219	79\\
220	79\\
221	79\\
222	79\\
223	79\\
224	79\\
225	79\\
226	79\\
227	79\\
228	79\\
229	79\\
230	79\\
231	79\\
232	79\\
233	79\\
234	79\\
235	79\\
236	79\\
237	79\\
238	79\\
239	79\\
240	79\\
241	79\\
242	79\\
243	79\\
244	79\\
245	79\\
246	79\\
247	79\\
248	79\\
249	79\\
250	79\\
251	79\\
252	79\\
253	79\\
254	79\\
255	79\\
256	79\\
257	79\\
258	79\\
259	79\\
260	79\\
261	79\\
262	79\\
263	79\\
264	79\\
265	79\\
266	79\\
267	79\\
268	79\\
269	79\\
270	79\\
271	79\\
272	79\\
273	79\\
274	79\\
275	79\\
276	79\\
277	79\\
278	79\\
279	79\\
280	79\\
281	79\\
282	79\\
283	79\\
284	79\\
285	79\\
286	79\\
287	79\\
288	79\\
289	79\\
};
\addplot [color=blue, line width=1.2pt, forget plot]
  table[row sep=crcr]{%
1	-20\\
2	-20\\
3	-20\\
4	-20\\
5	-20\\
6	-20\\
7	-20\\
8	-20\\
9	-20\\
10	-20\\
11	-20\\
12	-20\\
13	-20\\
14	-20\\
15	-20\\
16	-20\\
17	-20\\
18	-20\\
19	-20\\
20	-20\\
21	-20\\
22	-20\\
23	-20\\
24	-20\\
25	-20\\
26	-19\\
27	-18\\
28	-17\\
29	-16\\
30	-15\\
31	-15\\
32	-15\\
33	-15\\
34	-15\\
35	-15\\
36	-15\\
37	-15\\
38	-15\\
39	-15\\
40	-15\\
41	-15\\
42	-15\\
43	-15\\
44	-15\\
45	-15\\
46	-15\\
47	-15\\
48	-15\\
49	-15\\
50	-15\\
51	-15\\
52	-15\\
53	-15\\
54	-15\\
55	-15\\
56	-15\\
57	-15\\
58	-15\\
59	-15\\
60	-15\\
61	-15\\
62	-15\\
63	-15\\
64	-15\\
65	-15\\
66	-15\\
67	-15\\
68	-15\\
69	-15\\
70	-15\\
71	-15\\
72	-15\\
73	-15\\
74	-15\\
75	-15\\
76	-15\\
77	-15\\
78	-15\\
79	-15\\
80	-15\\
81	-15\\
82	-15\\
83	-15\\
84	-15\\
85	-15\\
86	-15\\
87	-15\\
88	-15\\
89	-15\\
90	-15\\
91	-15\\
92	-15\\
93	-15\\
94	-15\\
95	-15\\
96	-15\\
97	-15\\
98	-15\\
99	-15\\
100	-15\\
101	-15\\
102	-15\\
103	-15\\
104	-15\\
105	-15\\
106	-15\\
107	-15\\
108	-15\\
109	-15\\
110	-15\\
111	-15\\
112	-15\\
113	-15\\
114	-15\\
115	-15\\
116	-15\\
117	-15\\
118	-15\\
119	-15\\
120	-15\\
121	-15\\
122	-15\\
123	-15\\
124	-15\\
125	-15\\
126	-15\\
127	-14\\
128	-14\\
129	-14\\
130	-14\\
131	-12\\
132	-12\\
133	-12\\
134	-11\\
135	-11\\
136	-11\\
137	50\\
138	50\\
139	45\\
140	41\\
141	38\\
142	35\\
143	33\\
144	31\\
145	29\\
146	27\\
147	26\\
148	25\\
149	23\\
150	22\\
151	21\\
152	20\\
153	20\\
154	19\\
155	18\\
156	17\\
157	17\\
158	16\\
159	16\\
160	16\\
161	15\\
162	15\\
163	15\\
164	14\\
165	14\\
166	13\\
167	13\\
168	13\\
169	12\\
170	12\\
171	12\\
172	12\\
173	12\\
174	11\\
175	11\\
176	11\\
177	11\\
178	10\\
179	10\\
180	10\\
181	10\\
182	10\\
183	9\\
184	9\\
185	9\\
186	9\\
187	9\\
188	8\\
189	8\\
190	8\\
191	8\\
192	8\\
193	8\\
194	8\\
195	7\\
196	7\\
197	7\\
198	7\\
199	7\\
200	7\\
201	7\\
202	7\\
203	7\\
204	6\\
205	6\\
206	6\\
207	6\\
208	6\\
209	6\\
210	6\\
211	6\\
212	6\\
213	6\\
214	6\\
215	6\\
216	6\\
217	6\\
218	6\\
219	6\\
220	6\\
221	6\\
222	6\\
223	6\\
224	6\\
225	6\\
226	6\\
227	6\\
228	6\\
229	6\\
230	6\\
231	6\\
232	6\\
233	6\\
234	6\\
235	6\\
236	6\\
237	6\\
238	6\\
239	6\\
240	6\\
241	6\\
242	6\\
243	6\\
244	6\\
245	6\\
246	6\\
247	6\\
248	6\\
249	6\\
250	6\\
251	6\\
252	6\\
253	6\\
254	6\\
255	6\\
256	6\\
257	6\\
258	6\\
259	6\\
260	6\\
261	6\\
262	6\\
263	6\\
264	6\\
265	6\\
266	6\\
267	6\\
268	6\\
269	6\\
270	6\\
271	6\\
272	6\\
273	6\\
274	6\\
275	6\\
276	6\\
277	6\\
278	6\\
279	6\\
280	6\\
281	6\\
282	6\\
283	6\\
284	6\\
285	6\\
286	6\\
287	6\\
288	6\\
289	6\\
};
\end{axis}
\end{tikzpicture}%
}}
    %} 
    \end{minipage}
        \hfill
        \begin{minipage}[t]{0.2\textwidth}
        \subcaptionbox{Terrain roughness}{% 
       \includegraphics[width=1\linewidth]{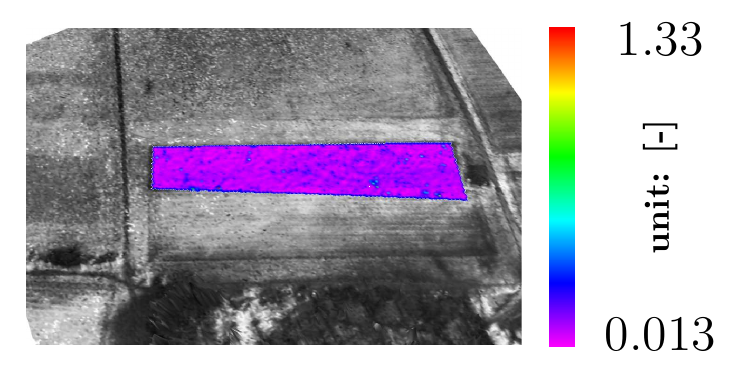}% 
    } 
    \end{minipage}
        \hfill
    \begin{minipage}[t]{0.2\textwidth}
    \subcaptionbox{Distance map\label{fig:rw_tobelhof_g}}{% 
       \includegraphics[width=1\linewidth]{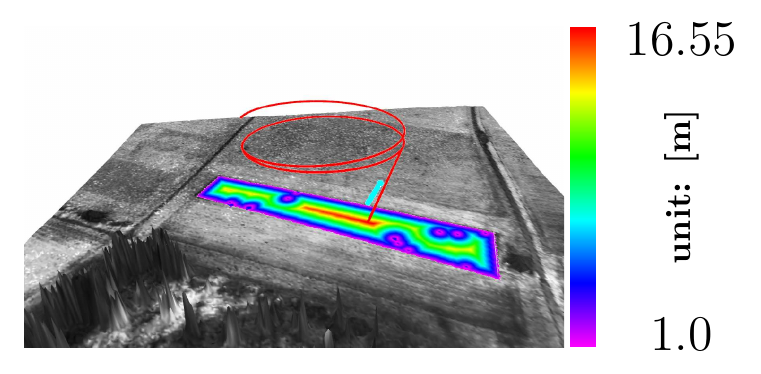}% 
    } 
    \end{minipage}
    \caption{
Manual flight with \emph{Techpod} in snowy scenery in Zurich, Switzerland. The employed camera is an \textit{IDS UI-3241LE} ($\unit[1.92]{MP}$). The experiment underlines the performance in a challenging environment and with an obliquely mounted camera.}
    \label{fig:rw_tobelhof} 
    \end{center}
    \vspace{-5pt}
  \end{figure*}

\reviews{The real-world experiments are analysed using datasets recorded onboard of \emph{AtlantikSolar} and \emph{Techpod}.}{review1:landing}{review1:real_time}
Details about the hardware setup and the employed platforms can be found in \cite{Oettershagen2017a} and \cite{Hinzmann2016}, respectively.
%Details about the hardware setup and the employed platforms AtlantikSolar and Techpod can be found in \cite{Oettershagen2017} and \cite{Hinzmann2016}, respectively.
%
%The results of the framework tested in real world scenarios are shown in Fig.~\ref{fig:rw_tobelhof} and   Fig.~\ref{fig:rw_brazil}.
%
The first experiment was conducted with the research platform \emph{Techpod} in snowy scenery in Switzerland (cf. Fig.~\ref{fig:rw_tobelhof}).
Fig.~\ref{fig:rw_tobelhof_b} and \ref{fig:rw_tobelhof_e} show the segmentation and classification of the landing region that received the highest score.
The ROI is then forwarded to the backend thread which generates the decision layers based on a dense point cloud.
The terrain slope and terrain roughness are used to compute the distance map (Fig. \ref{fig:rw_tobelhof_g}) which encodes the distance to the next hazard in form of a memory-friendly grid map.
From the score plots in Fig. \ref{fig:synth} and \ref{fig:rw_tobelhof} one can see that already the coarse grading can achieve a large separation between desired and undesired landing spots.
Depending on the UAV characteristics and desired landing spot, the final ROIs can be compared based on the output of the fine landing site evaluation. 
In the next experiment, the distance to terrain elevation during the landing approach is given as an example for such a fine ROI output statistic.

This second real-world experiment was conducted with \emph{AtlantikSolar} at the beach of Rio Par\'{a}, Brazil. 
Fig.~\ref{fig:rw_brazil_a} presents the overview mesh and camera poses for visualization, generated with \textit{Pix4D}.
Fig.~\ref{fig:rw_brazil_b} and \ref{fig:rw_brazil_c} show the $2.5$D elevation map, the estimated wind vector, and approach path to the selected landing site.
The touch down point in the landing site is selected based on the maximum distance to nearby hazards while considering obscuring obstacles.
The plots below show the path of the UAV with marked landing spot, wind speed measurements, and altitude profile during the approach path.
For instance, the margin between UAV altitude and terrain elevation is predicted to drop to approx. $\unit[2]{m}$,  $\unit[35]{m}$ before touch down.
\review{As discussed in Section \ref{sec:approach_vector}, the algorithm is designed to land against the wind vector, here with a magnitude of ca. $\unit[5.5]{m/s}$.
This has the advantage of reducing the aircraft's forward ground velocity, allowing for shorter landing ground distances and thus increasing the perceived descent angle with respect to \emph{ground}, yet still maintaining the chosen \emph{airmass-relative} flight path angle $\gamma_\text{land}$ (cf. Equation \ref{eq:approach_path}).}{review1:angle}
%
%\begin{figure*}[htb]
%%
%\centering \includegraphicsHighLowRes{width=\linewidth}{tikz/combined_tobelhof_with_plots_v2.pdf}{tikz/combined_tobelhof_with_plots_v2_compressed.png}{Manual flight with Techpod in snowy scenery in Zurich, Switzerland. The employed camera is an \textit{IDS UI-3241LE} ($\unit[1.92]{MP}$). The experiment underlines the performance in a challenging environment and with an obliquely mounted camera.}
%%
%\label{fig:rw_tobelhof}
%\end{figure*}
\setlength\figureheight{2.2cm}
\setlength\figurewidth{4.3cm}
%%%%%%%%%%%%%%%%%%%%%%%%%%%
  \begin{figure*}[htb] 
    \subcaptionbox{Overview mesh (\emph{Pix4D}), camera positions, and ROI\label{fig:rw_brazil_a}}{% 
      \includegraphics[width=0.4\linewidth]{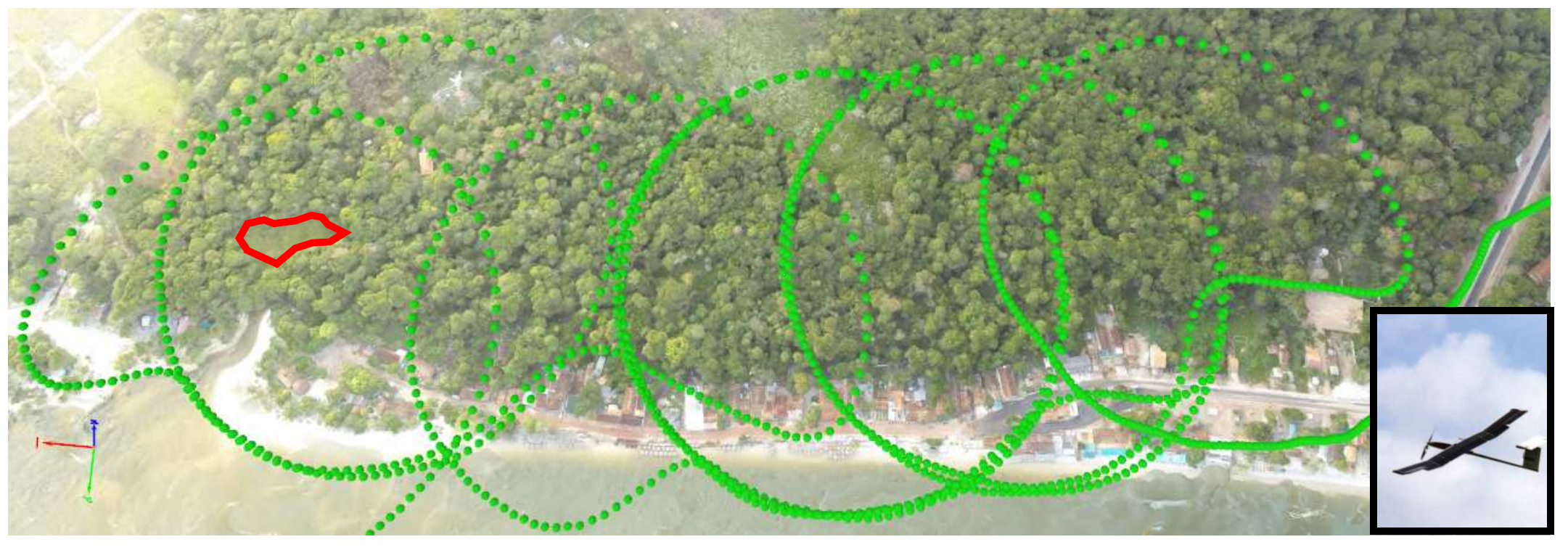}% 
    } 
    \hfill 
    \subcaptionbox{Approach path and wind vector\label{fig:rw_brazil_b}}{% 
      \includegraphics[width=0.25\linewidth, trim=0 100 0 0, clip=true]{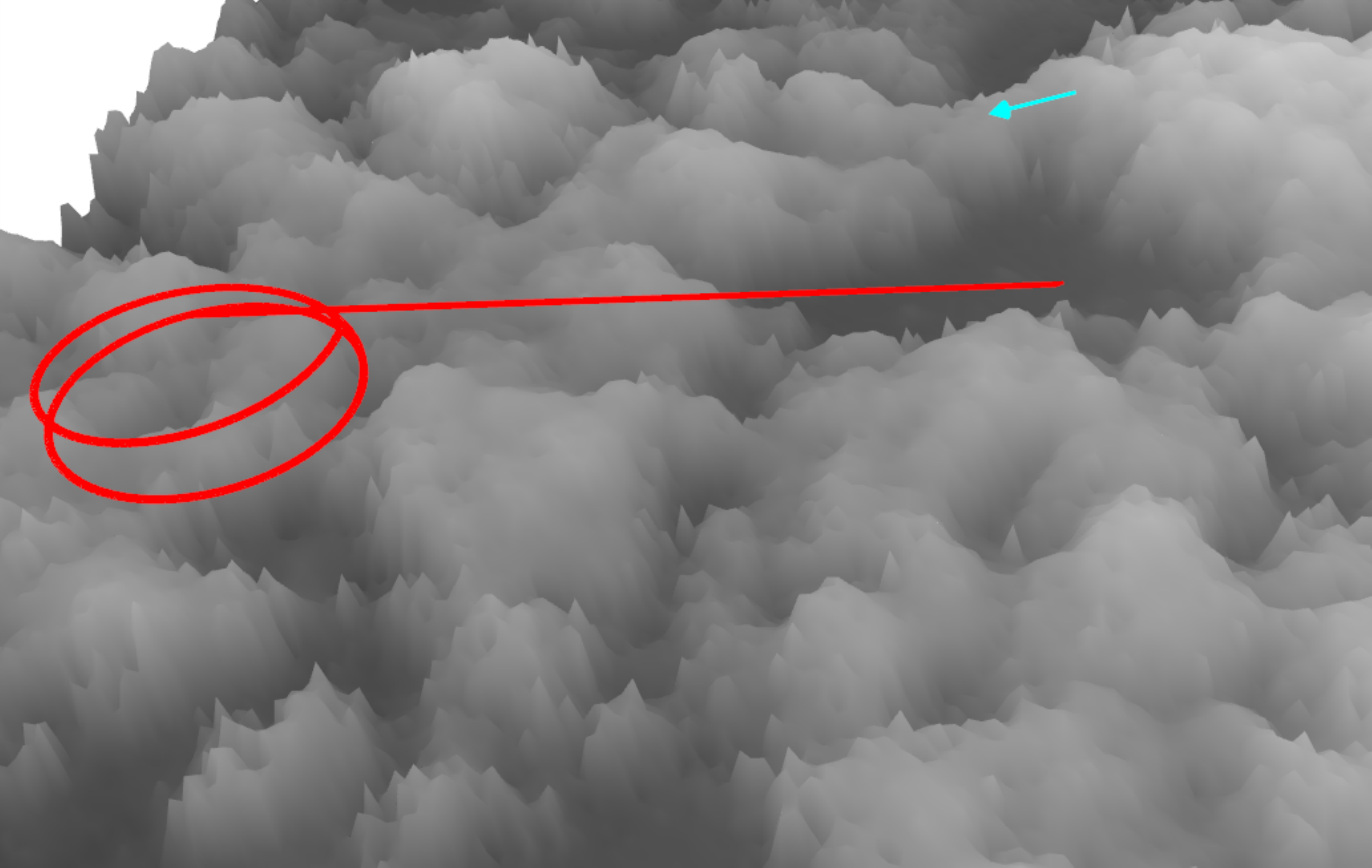}% 
    } 
	\hfill
    \subcaptionbox{Distance map\label{fig:rw_brazil_c}}{% 
      \includegraphics[width=0.27\linewidth]{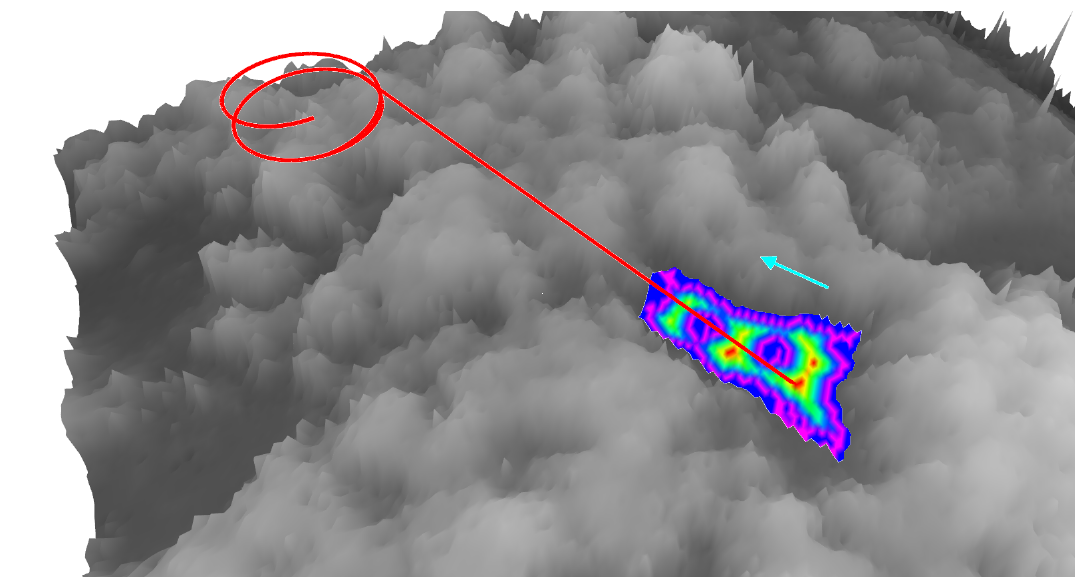}% 
    } 
    \\
    
    %\subcaptionbox{Altitude}{% 
      \raisebox{-10pt}{\scalebox{0.75}{\input{matlab/brazil/brazil_z.tikz}}}% 
    %} 
    	\hfill
       % \subcaptionbox{Path}{% 
      \raisebox{-10pt}{\scalebox{0.75}{\input{matlab/brazil/brazil_easting_northing.tikz}}}% 
    %} 
    	\hfill
       % \subcaptionbox{Wind}{% 
    \raisebox{-10pt}{\scalebox{0.75}{\input{matlab/brazil/brazil_wind.tikz}}}% 
    %} 
    	\hfill
    %    \subcaptionbox{Approach path}{% 
      \raisebox{-10pt}{\scalebox{0.75}{% This file was created by matlab2tikz.

\definecolor{mycolor1}{rgb}{0.00000,0.44700,0.74100}%
\definecolor{mycolor2}{rgb}{0.85000,0.32500,0.09800}%
\definecolor{mycolor3}{rgb}{0.92900,0.69400,0.12500}%
\begin{tikzpicture}

\begin{axis}[%
width=0.951\figurewidth,
height=\figureheight,
at={(0\figurewidth,0\figureheight)},
scale only axis,
xmin=0,
xmax=80,
xlabel style={font=\color{white!15!black}},
xlabel={Travelled distance [m]},
ymin=-2,
ymax=12,
ylabel style={font=\color{white!15!black}},
ylabel={Altitude [m]},
axis background/.style={fill=white},
xmajorgrids,
xminorgrids,
ymajorgrids,
legend image post style={scale=0.5},
legend columns=3, 
        legend style={
                    % the /tikz/ prefix is necessary here...
                    % otherwise, it might end-up with `/pgfplots/column 2`
                    % which is not what we want. compare pgfmanual.pdf
            /tikz/column 3/.style={
                column sep=1pt,
            },
        }
]
\addplot [color=mycolor1]
  table[row sep=crcr]{%
1	10.922\\
2	10.747\\
3	10.635\\
4	10.46\\
5	10.348\\
6	10.173\\
7	10.061\\
8	9.8861\\
9	9.7111\\
10	9.5992\\
11	9.424\\
12	9.3122\\
13	9.137\\
14	9.0253\\
15	8.85\\
16	8.675\\
17	8.5631\\
18	8.3879\\
19	8.2761\\
20	8.1009\\
21	7.9892\\
22	7.8139\\
23	7.7023\\
24	7.527\\
25	7.3519\\
26	7.24\\
27	7.0648\\
28	6.9531\\
29	6.7778\\
30	6.6662\\
31	6.4909\\
32	6.3158\\
33	6.2039\\
34	6.0288\\
35	5.917\\
36	5.7417\\
37	5.6302\\
38	5.4548\\
39	5.2798\\
40	5.1678\\
41	4.9927\\
42	4.8809\\
43	4.7057\\
44	4.5941\\
45	4.4187\\
46	4.2438\\
47	4.1317\\
48	3.9566\\
49	3.8448\\
50	3.6696\\
51	3.558\\
52	3.3826\\
53	3.2078\\
54	3.0956\\
55	2.9206\\
56	2.8087\\
57	2.6335\\
58	2.522\\
59	2.3465\\
60	2.2353\\
61	2.0595\\
62	1.8848\\
63	1.7726\\
64	1.5975\\
65	1.486\\
66	1.3104\\
67	1.1997\\
68	1.0234\\
69	0.84937\\
70	0.73664\\
71	0.56201\\
72	0.45064\\
73	0.27465\\
74	0.16903\\
75	-0.012717\\
};
\addlegendentry{$z_\text{plan}$}
\addplot [color=mycolor2]
  table[row sep=crcr]{%
1	3.2053\\
2	3.0542\\
3	1.9388\\
4	1.534\\
5	1.1573\\
6	2.2917\\
7	2.7037\\
8	2.7918\\
9	2.8706\\
10	3.6975\\
11	3.5358\\
12	3.9597\\
13	4.3977\\
14	4.7199\\
15	4.9107\\
16	5.0406\\
17	4.8984\\
18	4.7415\\
19	4.3957\\
20	4.0634\\
21	3.9938\\
22	3.5271\\
23	3.5961\\
24	3.2877\\
25	2.9541\\
26	2.7728\\
27	2.5991\\
28	2.4457\\
29	2.3087\\
30	1.7706\\
31	2.2889\\
32	2.2538\\
33	2.2828\\
34	3.0198\\
35	3.2213\\
36	3.381\\
37	3.038\\
38	2.5702\\
39	2.979\\
40	1.1981\\
41	2.5787\\
42	2.3931\\
43	2.1576\\
44	1.7529\\
45	1.23\\
46	0.74777\\
47	0.78567\\
48	0.43169\\
49	0.28442\\
50	0.026437\\
51	-0.21772\\
52	-0.25194\\
53	-0.33358\\
54	-0.31641\\
55	-0.27481\\
56	-0.27911\\
57	-0.24894\\
58	-0.19858\\
59	-0.16358\\
60	-0.26024\\
61	-0.1426\\
62	-0.13891\\
63	-0.13576\\
64	-0.18342\\
65	-0.12685\\
66	0.012822\\
67	-0.068181\\
68	-0.088922\\
69	-0.02635\\
70	0.014938\\
71	0.049218\\
72	0.0097113\\
73	0.064472\\
74	0.064348\\
75	-0.012717\\
};
\addlegendentry{$z_\text{elev}$}
\addplot [color=mycolor3]
  table[row sep=crcr]{%
1	7.7167\\
2	7.6928\\
3	8.6962\\
4	8.926\\
5	9.1907\\
6	7.8813\\
7	7.3573\\
8	7.0943\\
9	6.8405\\
10	5.9017\\
11	5.8882\\
12	5.3525\\
13	4.7393\\
14	4.3054\\
15	3.9393\\
16	3.6344\\
17	3.6647\\
18	3.6464\\
19	3.8804\\
20	4.0375\\
21	3.9954\\
22	4.2868\\
23	4.1062\\
24	4.2393\\
25	4.3978\\
26	4.4672\\
27	4.4657\\
28	4.5074\\
29	4.4691\\
30	4.8956\\
31	4.202\\
32	4.062\\
33	3.9211\\
34	3.009\\
35	2.6957\\
36	2.3607\\
37	2.5922\\
38	2.8846\\
39	2.3008\\
40	3.9697\\
41	2.414\\
42	2.4878\\
43	2.5481\\
44	2.8412\\
45	3.1887\\
46	3.49603\\
47	3.34603\\
48	3.52491\\
49	3.56038\\
50	3.643163\\
51	3.77572\\
52	3.63454\\
53	3.54138\\
54	3.41201\\
55	3.19541\\
56	3.08781\\
57	2.88244\\
58	2.72058\\
59	2.51008\\
60	2.49554\\
61	2.2021\\
62	2.02371\\
63	1.90836\\
64	1.78092\\
65	1.61285\\
66	1.297578\\
67	1.267881\\
68	1.112322\\
69	0.87572\\
70	0.721702\\
71	0.512792\\
72	0.4409287\\
73	0.210178\\
74	0.104682\\
75	0\\
};
\addlegendentry{$\Delta z$}
\end{axis}
\end{tikzpicture}%
}}% 
    %}
    	\hfill
    \caption[Caption for LOF]{Semi-automated flight at the beach of Rio Par\'{a}, Brazil, using \emph{AtlantikSolar} and a down-looking \textit{GoPro HERO3 Black} ($\unit[12]{MP}$): Takeoff and landing are performed in manual, the actual mission in automated GPS waypoint following mode. Subfig. (b) and (c) depict the $2.5$D elevation map: The darker the pixel, the lower the height or z-value of the elevation map's cell. The plots illustrate the incorporation of wind estimates into the LSD framework: By landing against the wind vector, the required landing distance is drastically reduced.} 
    \label{fig:rw_brazil}
  \end{figure*} 

%\begin{figure*}[htb]
%\includegraphicsHighLowRes{width=\linewidth}{tikz/combined_brazil_v3.pdf}{tikz/combined_brazil_v3.png}{Semi-automated\footnotemark flight at the beach of Rio Par\'{a}, Brazil, at sunrise using AtlantikSolar and a down-looking \textit{GoPro HERO3 Black} ($\unit[12]{MP}$). The figures illustrate the incorporation of wind estimates into the LSD framework. \review{Subfig. (b) and (c) depict the $2.5$D elevation map on gray scale. The darker the pixel, the lower the height or z-value of the elevation map's cell.}{review2:elevation_map}}
%% \setlength\figureheight{1.8cm}
%%\setlength\figurewidth{2.6cm}
%%\caption[] 
%\label{fig:rw_brazil}
%\end{figure*}
%\footnotetext{\review{}{review1:semi_automated}}
%

%
\section{Conclusion}
In this paper, we present a vision-based prior-free landing site detection algorithm which is designed for small UAVs, taking into account terrain texture, shape, roughness, and slope.
The wind field, which is estimated online, and obscuring obstacles are taken into consideration when computing a suitable landing spot while regarding UAV dynamics and safety margins.
To keep the problem complexity manageable, we segment the environment into regions and use a layered $2.5$D grid map for decision making.
%
%Furthermore, we combine a light-weight, real-time frontend and a backend which is updated based on the host's resources, with the modules running in different threads.
The implemented multi-threaded framework combines a light-weight, real-time frontend with a backend which is periodically updated based on the host's resources.
\review{The linear approach path, which is one output of our method, can be tracked as demonstrated in \cite{Oettershagen2017a}.
The actual landing attempt should furthermore be supported by a perception system, local re-planners and low-level autopilot logic to avoid previously unmapped or moving obstacles. 
In this paper, a simplistic key-frame selection algorithm was employed.
In a next step, an algorithm should be designed that guarantees complete coverage while minimizing the reconstruction uncertainty, utilized number of poses, and hence the computational costs.
Inter-matches and free-space carving \cite{Hornung2013} could be incorporated into the reconstruction process.}{review1:conclusion}
In future work, the classification and reconstruction uncertainty around a promising landing spot and approach path should be actively reduced by adapting the scanning pattern online and, in particular, by low-terrain flights to increase the ground resolution.
%
%Furthermore, a convolutional neural network could perform pixelwise segmentation as well as learn the features for classification instead of employing handcrafted features.
%

\vspace{-5pt}
\section*{Acknowledgment}
%ding from the Federal office armasuisse Science and Technology under project number
The research leading to these results has received funding from the \emph{Federal office armasuisse Science and Technology} under project number n\textdegree 050-45.
The authors wish to thank Felix Renaut for an initial implementation of the presented frontend, Lucas Teixeira (Vision for Robotics Lab, ETH Zurich) for sharing scripts that bridge the gap between \textit{Blender} and \textit{Gazebo}, and Philipp Oettershagen.
%\addtolength{\textheight}{-7pt}   
\balance

\bibliographystyle{ieeetr}
\bibliography{lib.bib}
\end{document}